\documentclass{article}

\usepackage[preprint]{neurips_2024}

\usepackage[utf8]{inputenc} 
\usepackage[T1]{fontenc}    
\usepackage{hyperref}       
\usepackage{url}            
\usepackage{booktabs}       
\usepackage{amsfonts}       
\usepackage{nicefrac}       
\usepackage{microtype}      
\usepackage{graphicx}  
\usepackage{float}
\usepackage{placeins}
\usepackage{booktabs}
\usepackage{array}
\usepackage{tabularx}
\usepackage{colortbl}
\usepackage{xcolor}
\usepackage{tcolorbox}   
\usepackage{xcolor}    
\usepackage{listings}   
\usepackage{enumitem}
\usepackage{tcolorbox}
\usepackage{pifont}  
\tcbuselibrary{breakable} 
\usepackage{listings}
\usepackage{bbm}
\usepackage{multirow}
\usepackage[normalem]{ulem}
\useunder{\uline}{\ul}{}

\title{Benchmark for Assessing Olfactory Perception of Large Language Models\thanks{Code and data: \url{https://github.com/Satarifard/Olfactory-Perception-benchmark}.}}

\author{%
  Eftychia~Makri \\
  Department of Electrical Engineering\\
  Yale University\\
  New Haven, CT, USA \\
  \And
  Nikolaos~Nakis \\
 Human Nature Lab\\
  Yale University\\
  New Haven, CT, USA \\
  \And
  Laura~Sisson\\
  Patina\\
  New York, NY, USA \\
  \And
  Gigi~Minsky\\
  Department of Ecology \\ University of California\\
  Santa Barbara, CA, USA \\
  \And
  Leandros~Tassiulas \\
  Department of Electrical Engineering\\
  Yale University\\
  New Haven, CT, USA \\
  \And
  Vahid~Satarifard \\
  Human Nature Lab\\
  Yale University\\
  New Haven, CT, USA \\
  \texttt{vahid.satarifard@yale.edu} \\
  \And
  Nicholas A. Christakis \\
  Human Nature Lab\\
  Yale University\\
  New Haven, CT, USA \\
  \texttt{nicholas.christakis@yale.edu} \\
}

\begin{document}

\maketitle

\begin{abstract}
Here we introduce the Olfactory Perception (OP) benchmark, designed to assess the capability of large language models (LLMs) to reason about smell. The benchmark contains 1,010 questions across eight task categories spanning odor classification, odor primary descriptor identification, intensity and pleasantness judgments, multi-descriptor prediction, mixture similarity, olfactory receptor activation, and smell identification from real-world odor sources. Each question is presented in two prompt formats, compound names and isomeric SMILES, to evaluate the effect of molecular representations. Evaluating 21 model configurations across major model families, we find that compound-name prompts consistently outperform isomeric SMILES, with gains ranging from +2.4 to +18.9 percentage points (mean $\approx$ +7 points), suggesting current LLMs access olfactory knowledge primarily through lexical associations rather than structural molecular reasoning. The best-performing model reaches 64.4\% overall accuracy, which highlights both emerging capabilities and substantial remaining gaps in olfactory reasoning. We further evaluate a subset of the OP across 21 languages and find that aggregating predictions across languages improves olfactory prediction, with AUROC\,=\,0.86 for the best performing language ensemble model. LLMs should be able to handle olfactory and not just visual or aural information.
\end{abstract}

\section{Introduction}

The sense of smell occupies a unique position among human perceptual modalities \citep{mainland,secundo}. Unlike vision, where wavelengths map predictably to colors \citep{Stockman2000TheSS}, or audition, where frequencies correspond to pitch \citep{oxen}, olfaction operates through a complex interplay between molecular structure and receptor biology that remains incompletely understood \citep{mainland,Buck1991ANM,bushdid2014humans}. Humans can distinguish over a trillion distinct odors \cite{bushdid2014humans,erratum}, though estimates are still debated \citep{10.7554/eLife.08127}, yet predicting how a molecule will smell from its chemical structure alone has long eluded computational approaches \citep{keller2017predicting}. This structure-odor relationship represents one of the most challenging frontiers in sensory science \cite{keller2017predicting, lee2023principal,secundo}.

Large language models have demonstrated remarkable capabilities across diverse domains, from mathematical reasoning to code generation \cite{brown2020languagemodelsfewshotlearners,chen2021evaluatinglargelanguagemodels}. Recent work has shown that these models, particularly those trained with explicit reasoning objectives \citep{han2024generalistspecialistsurveylarge}, can perform sophisticated tasks in chemistry, interpreting SMILES (Simplified Molecular Input Line Entry System) \citep{Weininger1988SMILESAC} strings, predicting reaction products, and even elucidating molecular structures from NMR spectra \cite{runcie2025assessing, mirza2025framework}. These advances raise a natural question: can LLMs bridge the gap between molecular structure and sensory perception? Specifically, do these models possess knowledge about how molecules smell?

This question carries both scientific and practical significance. From a scientific perspective, understanding what LLMs have learned about olfaction provides insight into how chemical knowledge is encoded in these systems. From an applied standpoint, LLMs capable of olfactory reasoning could accelerate fragrance design \cite{mao}, flavor development \cite{Ge2025MachineLF}, and the identification of malodorous contaminants in consumer products \cite{Bartsch2016AnalysisOO}. Yet despite growing interest in evaluating LLM sensory alignment across modalities including color and auditory perception, as well as taste \cite{marjieh2024large}, olfaction has remained notably absent from systematic evaluation.

Prior work exploring the intersection of AI and olfaction has taken different approaches. SNIFF AI \cite{zhong2024sniff} investigated human-AI perceptual alignment through user studies where participants described scents and embedding models attempted identification, revealing limited alignment, with only a 27.5\% success rates. Other studies have examined whether LLMs can recover olfactory-semantic relationships from natural language \cite{kurfali2025representations}, finding that models like GPT-4o capture similarity judgments derived from word-based assessments. Meanwhile, specialized machine learning systems using graph neural networks have achieved human-level odor prediction from molecular structure \cite{keller2017predicting,lee2023principal}. However, no existing work systematically evaluates whether general-purpose LLMs possess factual knowledge about olfactory properties through structured question-answering with objective ground-truth answers.

We address this gap by introducing the Olfactory Perception (OP) benchmark, a comprehensive evaluation framework comprising 1,010 questions across eight task categories. Our benchmark spans the full complexity of olfactory perception: from basic odor detection \cite{mayhew2022transport} and classification to nuanced judgments of intensity and pleasantness \cite{keller2017predicting}, from single-molecule descriptor identification \cite{FIG} to the perceptual similarity of complex mixtures \cite{bushdid2014humans,snitz2013predicting,  ravia2020measure,satarifard2025high}, and from semantic odor labeling \cite{lee2023principal} to the biological mechanisms of receptor activation \cite{lalis2024m2or}. Ground-truth answers are derived from established datasets in olfactory science, ensuring that our evaluation reflects genuine knowledge rather than superficial pattern matching.

Odor language varies substantially across cultures and linguistic communities. Some languages support more abstract, dedicated olfactory vocabularies and enable more efficient odor naming \cite{MAJID2014266,MAJID2018}. Motivated by this cross-linguistic diversity and evidence that human olfaction sits at the intersection of language, culture, and biology \cite{MAJID2021111}, we evaluate a subset of our benchmark across multiple languages to test whether LLM olfactory knowledge is robust to linguistic framing rather than specific to English. We additionally report a cross-lingual majority-vote setting, since different languages may provide more informative descriptor inventories for different odors.

A distinctive feature of our approach is the dual-prompting strategy employed throughout the benchmark. Each question is presented in two forms: one using isomeric SMILES molecular notation and another using common compound names. Because odor perception is stereospecific, different isomers, especially enantiomers, can smell very different; for example, (R)-carvone is perceived as spearmint-like whereas (S)-carvone is perceived as caraway-like. This design enables direct comparison of how different molecular representations affect olfactory reasoning, an experimental dimension unexplored in prior work. Similar methodology has proven informative in chemistry benchmarks, where canonical versus randomized SMILES representations yield different model performance \cite{runcie2025assessing}. The contrast between these conditions assess whether models are genuinely reasoning about molecular properties or merely retrieving associations from training data.

Our evaluation encompasses both reasoning and non-reasoning models across multiple providers, including OpenAI's GPT and o3 series, Google's Gemini, Anthropic's Claude, Meta's Llama, xAI's Grok, and DeepSeek. We systematically vary reasoning budgets where applicable, enabling analysis of how extended deliberation affects olfactory task performance, an approach that has revealed substantial performance differences in chemistry tasks \cite{runcie2025assessing, mirza2025framework}. Our findings reveal that while current models achieve moderate success on certain tasks, substantial gaps remain, particularly for questions requiring genuine molecular reasoning rather than factual recall about well-known compounds.


The contributions of this work are threefold. First, we introduce a comprehensive benchmark for evaluating LLMs on olfactory reasoning, comprising 1{,}010 questions across eight task categories grounded in peer-reviewed olfactory science. Second, we establish baseline performance across state-of-the-art models (spanning multiple providers and reasoning configurations), identifying both emerging capabilities and systematic failures. Third, via a dual-prompting methodology (compound names vs. isomeric\ SMILES), we provide new insight into how molecular representation format shapes model performance on perceptual prediction tasks, helping to distinguish structural reasoning from lexical association. Together, these contributions lay the groundwork for future research at the intersection of artificial intelligence and olfactory science.

\section{Related work}
\textbf{LLM benchmarks for Chemistry and Molecular Understanding.} Recent benchmarks have evaluated the chemical reasoning capabilities of large language models. ChemBench \cite{mirza2025framework} established a comprehensive chemistry benchmark with over 2,700 questions across eight sub-categories including organic chemistry, toxicity, and medicinal chemistry, finding that leading models outperform human chemists on average. ChemIQ \cite{runcie2025assessing} focuses specifically on molecular comprehension with 816 questions testing SMILES interpretation, atom counting, reaction prediction, and NMR structure elucidation-reasoning models (o3-mini, Gemini Pro 2.5); it achieved 50-57\% accuracy, substantially outperforming non-reasoning models (3-7\%). Additionally, LlaSMol \cite{yu2024llasmol} introduced SMolInstruct, a large-scale instruction tuning dataset with over 3 million samples across 14 chemistry tasks, and demonstrated that fine-tuned open-source LLMs can substantially outperform GPT-4 on chemistry tasks, with SMILES representations outperforming SELFIES for molecular understanding. Other benchmarks include GPQA \cite{rein2024gpqa} with expert-written chemistry questions, MMLU \cite{hendrycks2021measuring} which includes chemistry among 
its 57 subjects, and a comprehensive eight-task benchmark \cite{guo2023can}.

While these benchmarks comprehensively assess structural and general chemical understanding, none evaluate whether LLMs can reason about the perceptual properties of molecules, in particular how they smell. Our work addresses this gap by testing the translation from molecular structure to olfactory perception.

\textbf{LLMs and Sensory Perception.} A growing body of work investigates whether LLMs align with human sensory judgments. It has been shown that GPT models produce similarity judgments significantly correlated with human data across six modalities: pitch, loudness, colors, consonants, taste, and timbre \cite{marjieh2024large}. Notably, olfaction was excluded from their evaluation, leaving an open question about LLM capabilities in this domain. This line of inquiry is extended \cite{xu2025large} by comparing LLM representations of ${\sim}$4,442 lexical concepts against human norms across non-sensorimotor, sensory, and motor domains, finding that alignment decreases systematically from non-sensorimotor to sensory domains and is minimal for motor concepts; adding visual training improves sensory but not motor alignment. These findings highlights a grounding gap that is particularly acute for embodied perception. Even within well-studied modalities, limitations persist. For example, GPT-3 through GPT-4o were tested on color-word associations against over 10,000 Japanese participants \cite{fukushima2025advancements} and found that GPT-4o peaked at approximately 50\% accuracy, with strong variation across word categories. Together, these findings suggest that LLMs capture substantial but incomplete perceptual structure from text, with performance degrading for modalities that are less frequently described in language. Olfaction, rarely discussed in explicit perceptual terms, represents a natural test case for these limitations.

\textbf{LLMs and Olfaction.} Concurrent work has begun exploring the olfaction gap specifically. SNIFF AI \cite{zhong2024sniff} investigated human-AI perceptual alignment for smell through user studies where participants described scents and an LLM embedding model attempted identification. Their findings revealed limited alignment, with biases toward certain scents (e.g., lemon, peppermint) and systematic failures on others (e.g., rosemary), achieving only 27.5\% success in their scent description task.

Most recently, three generations of language models are systematically evaluated \cite{kurfali2025representations}, from static word embeddings (Word2Vec, FastText) to encoder-based models (BERT) and decoder-based LLMs (GPT-4o, Llama 3.1), on their ability to recover olfactory information from natural language. Testing under nearly 200 training configurations across three odor datasets, they found that GPT-4o excels at simulating olfactory-semantic relationships, particularly on tasks where odor similarities are derived from word-based assessments. However, their evaluation focuses on semantic similarity judgments rather than factual accuracy about olfactory properties.

Our benchmark complements these approaches by testing whether LLMs possess correct factual knowledge about odor classification, perceptual attributes, and biological mechanisms through structured question-answering with objective ground-truth answers.

\textbf{Machine Learning for Olfactory Perception.} Parallel to LLM development, specialized machine learning models have made significant progress in computational olfaction. Two DREAM Olfaction Prediction Challenges \cite{keller2017predicting,satarifard2025high} established foundational work on predicting semantic descriptors of single molecules and olfactory perceptual similarity of mixtures. In addition,  a Principal Odor Map has been introduced \cite{lee2023principal} using graph neural networks, and it achieved human-level proficiency in describing odor qualities across 500,000 potential scent molecules. More recently, Mol-PECO \cite{zhang2023mol} extended graph neural network approaches with Coulomb matrix representations to predict 118 odor descriptors from molecular structure, achieving AUROC of 0.813. POMMIX \cite{tom2025molecules} further extended the POM to olfactory mixture similarity, combining attention-based mixture representations with inductive biases to reach Pearson $\rho = 0.779$ on human perceptual similarity data, directly relevant to the mixture-level tasks we evaluate in this benchmark.

These specialized systems demonstrate that machine learning can capture structure-odor relationships when purpose-built for olfactory tasks. However, they do not address whether general-purpose language models that are primarily trained on text without explicit olfactory supervision possess latent knowledge about smell.

\textbf {Positioning of Our Work.} Our Olfactory Perception (OP) benchmark is, to our knowledge, the first comprehensive 
evaluation of LLM olfactory reasoning through structured factual question-answering. 
While ChemIQ and ChemBench focus on molecular structure understanding, and SNIFF AI 
evaluates embedding-space alignment through subjective descriptions, our benchmark 
tests discrete factual knowledge across eight task categories: odor classification, 
odor primary descriptors, intensity, pleasantness, multi-descriptor identification, 
mixture similarity, receptor activation, and smell identification. Ground-truth answers 
are derived from established olfactory science datasets and resources 
\citep{mayhew2022transport, keller2017predicting, lee2023principal, lalis2024m2or, 
snitz2013predicting, bushdid2014humans, ravia2020measure, FIG, kreissl2022odorant}. 
Furthermore, our dual-prompting strategy (isomeric SMILES vs.\ compound names) enables direct 
comparison of how different molecular representations affect olfactory reasoning-an 
experimental design not explored in prior work. Table~\ref{tab:related-work} summarizes the positioning of our benchmark relative to prior work.

\begin{table}[!hb]
\centering
\caption{Comparison of related work on LLMs, chemistry, and olfaction.}
\label{tab:related-work}
\small
\renewcommand{\arraystretch}{1.2}
\begin{tabular}{p{2.8cm}p{2.2cm}p{4.5cm}p{4cm}}
\toprule
\textbf{Work} & \textbf{Input} & \textbf{Specific Tasks} & \textbf{What is Tested} \\
\midrule
\multicolumn{4}{l}{\textbf{Chemistry LLM benchmarks}} \\
\midrule
ChemBench \cite{mirza2025framework} & Chemistry questions & General, organic, analytical, toxicity chemistry (2,700 Q) & Structural \& reaction knowledge \\
ChemIQ \cite{runcie2025assessing} & SMILES & Carbon counting, ring counting, SMILES-to-IUPAC, NMR elucidation, reaction prediction (816 Q) & Molecular comprehension \\
LlaSMol \cite{yu2024llasmol} & SMILES/SELFIES & Name conversion, property prediction, synthesis (14 tasks, 3M samples) & Fine-tuned LLM chemistry \\
\midrule
\midrule
\multicolumn{4}{l}{\textbf{LLMs and Sensory Perception}} \\
\midrule
Marjieh et al.\ \cite{marjieh2024large} & Sensory stimuli & Pairwise similarity judgments across pitch, loudness, color, consonants, taste, timbre & 6 modalities (olfaction excluded) \\
Xu et al.\ \cite{xu2025large} & Lexical concepts & Conceptual similarity across sensorimotor and non-sensorimotor domains ($\sim$4,442 concepts) & Grounding gap in sensory/motor domains \\
Fukushima et al.\ \cite{fukushima2025advancements} & Color stimuli & Color-word associations (17 colors $\times$ 80 words, 10K+ human baseline) & Color perception alignment \\
\midrule
\midrule
\multicolumn{4}{l}{\textbf{LLMs and Olfaction}} \\
\midrule
SNIFF AI \cite{zhong2024sniff} & Human descriptions & Human describes scent, LLM identifies source (27.5\% success) & Description-to-scent mapping \\
Kurfal{\i} et al.\ \cite{kurfali2025representations} & Odor word pairs & Pairwise similarity ratings across 3 odor datasets & Semantic similarity of odor words \\
Kurfal{\i} et al.\ \cite{kurfali2023enhancing} & Image + text & Detect if image and text share olfactory source & Multimodal matching \\
Esteban-Romero et al.\ \cite{esteban2025synthesizing} & Image + text & Image-text smell matching (F1=0.76) & Multimodal matching \\
\midrule
\midrule
\multicolumn{4}{l}{\textbf{ML for Olfactory Perception (non-LLM)}} \\
\midrule
First DREAM Challenge \cite{keller2017predicting} & Molecular features & Predict intensity, pleasantness, 19 semantic descriptors (476 odorants) & Specialized ML models \\
Second DREAM Challenge \cite{satarifard2025high} & Molecular features & Predict olfactory perceptual distance of mixtures & Specialized ML models \\
Principal Odor Map \cite{lee2023principal} & Molecular graphs & Predict odor qualities, similarity (500K molecules) & GNN on molecules \\
Mol-PECO \cite{zhang2023mol} & Molecular graphs (Coulomb matrix) & Predict 118 odor descriptors (8,503 molecules) & Deep learning for QSOR \\
POMMIX \cite{tom2025molecules} 
& Molecular graphs (GNN + attention) 
& Predict olfactory mixture similarity 
& Mixture representation learning \\
\midrule
\midrule
\textbf{OP benchmark (Ours)} & \textbf{isomeric SMILES-Compound Name} & \textbf{Odor classification, primary descriptor, intensity, pleasantness, rate-all-that-apply, mixture similarity, receptor activation, smell identification (1,010 Q)} & \textbf{LLM olfactory knowledge} \\
\bottomrule
\end{tabular}
\end{table}

\section{Olfactory Perception Benchmark}

\begin{figure}[!t]
  \centering
  \includegraphics[width=\linewidth]{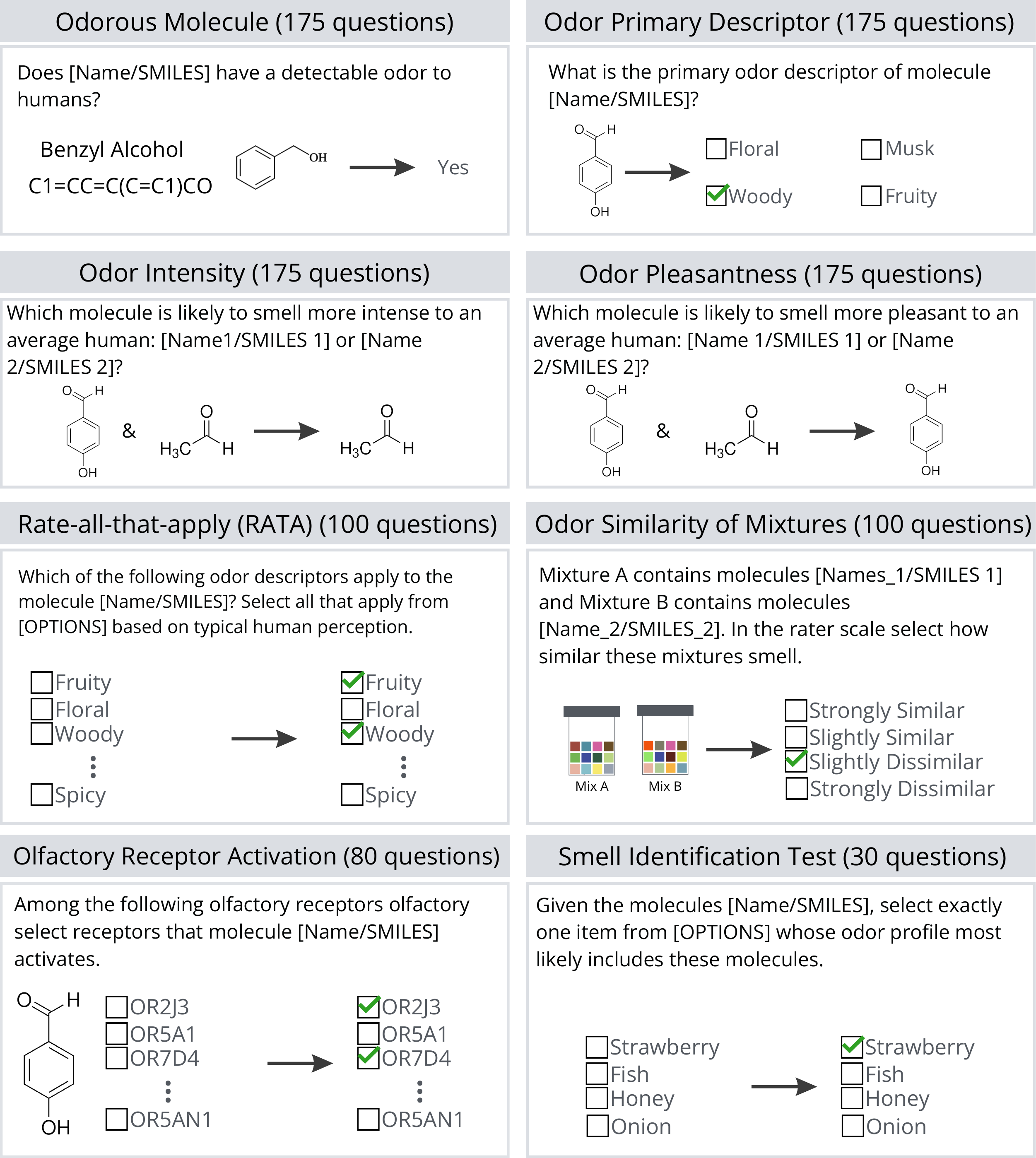}
  \caption{Olfactory Perception (OP) benchmark task description and output format across tasks.}
  \label{fig:figure1}
\end{figure}

We introduce a unified benchmark of 1,010 olfaction questions spanning odor detectability, semantic odor description, perceptual judgments, mixture similarity, receptor activation, and smell identification test from mixtures. Figure~\ref{fig:figure1} provides an overview of the benchmark, while Table~\ref{tab:oi-benchmark-categories} summarizes the eight question categories, their data sources, and the number of questions per category. Each item is presented in two equivalent formats: i) isomeric SMILES (prompt 1), and ii) compound names (prompt 2), to separate structure-based reasoning from name priors. Tasks are multiple choice from constrained lists of options to enable consistent automatic scoring across models.
\small
\begin{table}[t]
  \caption{Question categories and data sources used in the benchmark.}
  \label{tab:oi-benchmark-categories}
  \centering
  \begin{tabular}{p{2.7cm}p{2.5cm}p{0.5cm}p{4.2cm}p{0.8cm}p{1cm}}
    \toprule
    Question Category & Options  & n & Note & Type & Reference \\
    \midrule
    Odor Classification &
    Odorous/Odorless &
    175 &
    Molecular Weight $\le$ 350.0 50/50 &
    Pure &
    \cite{mayhew2022transport} \\
    \midrule
    Odor Descriptor &
    1 answer, 3 distractors &
    175 &
    Stratified to represent less dominant descriptors, 29 Descriptors total &
    Pure &
    \cite{FIG} \\
    \midrule
    Odor Intensity &
    High/Low &
    175 &
    Above and below median, with score difference of 10 &
    Pure &
    \cite{keller2017predicting} \\
    \midrule
    Odor Pleasantness &
    High/Low &
    175 &
    Above and below median, with score difference of 10 &
    Pure &
    \cite{keller2017predicting} \\
    \midrule
    Rate-all-that-apply &
    2-5 answers from 138 distractors &
    100 &
    25 questions for 2,3,4, and 5 descriptors &
    Pure &
    \cite{lee2023principal} \\
    \midrule
    Odor Similarity of Mixtures &
    Rater scale &
    100 &
    Mixtures with 2-10 molecules &
    Mixture &
    \cite{snitz2013predicting,bushdid2014humans,ravia2020measure}\\
    \midrule
    Olfactory Receptor Activation &
    1-10 answers from 4-10 ORs &
    80 &
    Human OR, 4-10 OR studied &
    Pure &
    \cite{lalis2024m2or} \\
    \midrule
    Smell Identification Test &
    1 answer, 3 distractors &
    30 &
    &
    Mixture &
   \cite{kreissl2022odorant} \\
    \bottomrule
  \end{tabular}
\end{table}

\subsection{Odor Classification (OC)}
Odor classification is a binary task that asks whether a given molecule is \textit{Odorous} or \textit{Odorless} (175 questions, \%50 Odorous). This task assess basic odor detectability prediction from chemical identity of a molecule. Molecules were obtained from previously curated dataset \cite{mayhew2022transport} and molecular weight was constrained to be  $\mathrm{\leq 350.0}$ (so as to be in the smellable range); here were presented to LLMs in random order. We consider odor classification to be a simple task.

\subsection{Odor Primary Descriptor (OPD)}
Odor primary descriptor is a multi-class task where the model selects the single descriptor for a molecule from a provided list of four options (175 questions). This task evaluates whether models can map chemicals to main semantic odor categories. The molecules were obtained from the 2020 version of the IFRA fragrance ingredient glossary (FIG) \cite{FIG}. Besides Primary descriptor, FIG provides secondary and tertiary descriptors, we excluded these descriptors from the list of distractors to obtain a more robust evaluation. Selection of molecules was done on a stratified set of descriptors to represent less dominant descriptors and a total of 29 descriptors are present among questions. We consider odor primary pescriptor as a simple task.

\subsection{Odor Intensity and Pleasantness (OIn and OPl)}

To evaluate model assessment of two olfactory dimensions (\textit{i.e.,} intensity and pleasantness), we use two paired-comparison tasks, where the model chooses which of two molecules is more intense or more pleasant (175 molecular pairs for intensity, and 175 molecular pairs for pleasantness comparisons). Ground truth data are obtained from prior work \cite{keller2017predicting} which provides mean human subjective rating ranging a 0-100 scale. Molecular pairs were selected to be above and below median with a minimum score difference of 10. Furthermore, prompts asks model to rate the intensity and pleasantness in a 0–100 scale. We consider odor intensity and pleasantness category to be a simple tasks.

\subsection{Rate-all-that-apply (RATA)}
To assess more complex olfactory perceptual capability, we employ a multilabel semantic profiling task (100 questions), where, given a molecule and a descriptor lexicon (138 odor descriptors), the model selects all descriptors that apply in describing the odor of the molecule. We selected 100 molecules from an integrated dataset of the Good Scents and Leffingwell Associates (GS-LF) \cite{lee2023principal}.  We constrained our selection to molecules with 2 to 5 answers from 138 descriptors, with 25 questions for 2,3,4, and 5 descriptors, respectively. We consider RATA categorization as an intermediate hard task.

\subsection{Odor Similarity of Mixtures (OS)}
In odor similarity of mixtures, we evaluate a categorical mixture-perception task (100 questions) where the model compares two mixtures (each mixture is a set of 2-10 molecules) and predicts an ordinal similarity label (e.g., from strongly similar to strongly dissimilar). We selected 100 mixture pairs from various datasets \cite{snitz2013predicting,bushdid2014humans,ravia2020measure} and after standardization of perceptual distance, similarity values were categorized into four bins of a categorical rating scale, including:  strongly similar, slightly similar, slightly dissimilar, and strongly dissimilar. This task deals with mixture-level olfactory perception, and we classify it as intermediate hardship.

\subsection{Olfactory Receptor Activation (ORA)}
In this task, we aim to assess model capability in identifying olfactory receptor (OR) activation for pure molecules. We use a multilabel task (80 questions) that asks which receptors (from a candidate set of 4-10 human OR gene IDs) are activated by a given molecule. The molecules-OR pairs were obtained from the M2OR dataset \cite{lalis2024m2or}. We constrained our selection of molecules-OR pairs to cases where 4 to 10 ORs activation were experimentally studied, and 1 to 10 ORs were observed to be activated. We consider ORA category to be a hard task.

\subsection{Smell Identification Test (SIT)}
We curated a multiple-choice smell identification task (30 questions) with molecular mixture information. The model selects the most likely odor source from a four-option set given a molecular profile. We obtained the list of volatile organic compounds for different food items from the Leibniz-lsb\@tum odorant database \cite{kreissl2022odorant}. The items studied includes the odor profile for mango, peanut, hazelnut, tomato, apple, walnut, raspberry, peach, honey, parsley, grapefruit, pineapple, strawberry, apricot, rice, grape, popcorn, orange, cheese, melon, leather, chocolate, coffee, onion, fish, beer, whisky, red wine, prawn, and bread. We consider the SIT category to be a hard task.

\subsection{Multilingual translation}
We generated a multilingual versions of the RATA tasks by translating the English odor-descriptor vocabulary and the UI/instruction text into each target language using \texttt{GPT-5.2}. We prompted the model to translate in a perfumery/sensory context and to produce brief, natural descriptors; we used high-reasoning settings for disambiguation. Descriptor translations were normalized to lowercase for consistency, while UI strings were translated separately to preserve natural phrasing.

\subsubsection{Quality control and fallbacks}
To reduce hallucinated or awkward terms, we applied a lightweight two-step check for any uncertain cases: the model proposes candidate single-word descriptors and then selects the best option based on whether it is a real word in the target language and whether it fits olfactory usage. If no suitable translation is found, we fall back to the closest candidate; if generation fails entirely, we retain the original English descriptor. For languages where compounds are standard (e.g., German, Swedish), compound forms were allowed as long as they remained whitespace-free.

\section{Experiments}
\label{sec:experiments}

To comprehensively evaluate olfactory reasoning capabilities across the current LLM landscape, 
we select models spanning multiple providers, architectures, and reasoning configurations. 
A key dimension of our evaluation is the effect of reasoning budget on olfactory task 
performance, motivated by findings in chemistry benchmarks showing that extended reasoning 
substantially improves molecular understanding~\cite{runcie2025assessing}. 
We organize our evaluation into two groups:

\begin{itemize}[leftmargin=*]
\item \textbf{Closed-Source Models:} We evaluate models from five providers. 
From OpenAI \cite{achiam2023gpt}, we test the reasoning models o3 (high) and o4-mini (high), 
the GPT-5 family \cite{singh2025openai} at two reasoning levels (low, high), GPT-5 Pro, GPT-5.2 Pro, 
and GPT-OSS 120B\cite{agarwal2025gpt} at high reasoning settings. 
From Google, we evaluate Gemini 2.5 Pro~\cite{comanici2025gemini} at three reasoning 
budgets (8K, 16K, and 32K tokens) to isolate the effect of reasoning depth. 
From xAI, we include Grok 3 Mini at low and high reasoning and Grok 4.1 Fast. 
From Anthropic, we evaluate Claude Sonnet 4.5, Claude Opus 4.5~\cite{anthropic2025opus45}, 
and Claude Opus 4.6 ~\cite{anthropic2026opus46} at two reasoning budgets (high, max).

\item \textbf{Open-Source Models:} We evaluate 
DeepSeek Reasoner~\cite{guo2025deepseek} at three reasoning budgets 
(8K, 16K, and 32K tokens) and 
Llama-3.3-70B-Instruct~\cite{dubey2024llama}. 
These models provide a baseline for open-weight performance on olfactory tasks.
\end{itemize}

\noindent In total, we evaluate 21 model configurations across 6 providers. All models were queried through their respective provider  APIs without enabling web search, tool use, or any 
external retrieval capabilities; by default, none of the  APIs grant models access to external resources during  inference. All models are prompted identically using both isomeric SMILES and compound name 
representations; full prompt templates and worked examples are provided 
in Appendix~\ref{appendix:prompts}. For reasoning models, we systematically vary the reasoning 
budget to analyze the relationship between computational deliberation and 
olfactory task accuracy.

\textbf{Evaluation Metrics.} We adopt task-appropriate metrics to capture performance across the
diverse question formats in the OP~benchmark.  For single-answer tasks
(OC, OPD, OIn, OPl, OS, SIT), we report \emph{any-overlap accuracy}: a
question is scored correct if the set of tokens extracted from the
model's response intersects the ground-truth token set.  For
multi-answer tasks (RATA, ORA), we employ \emph{per-question multilabel
F1}, which provides partial credit when models correctly identify a
subset of applicable labels and penalises both spurious and missing
predictions.  For the three continuous-rating tasks (OI, OPl, OS), we
additionally compute Pearson correlations between predicted and human
psychophysical ratings.  Overall accuracy is the unweighted arithmetic
mean of the eight per-task scores. More details on the answer extraction pipeline are provided in Appendix ~\ref{appendix:answer_extraction}.

\begin{table*}[t]
\centering
\small
\setlength{\tabcolsep}{3pt}
\caption{Olfactory Perception (OP) benchmark results using compound name prompts. Accuracy (\%) is reported per task and overall (unweighted mean across tasks). Abbreviations: \textbf{OC} = Odor Classification ($n$=175), \textbf{OPD} = Odor Primary Descriptor ($n$=175), \textbf{OIn} = Odor Intensity ($n$=175), \textbf{OPl} = Odor Pleasantness ($n$=175), \textbf{RATA} = Rate-All-That-Apply ($n$=100), \textbf{OS} = Odor Similarity ($n$=100), \textbf{ORA} = Olfactory Receptor Activation ($n$=80), \textbf{SIT} = Smell Identification Test ($n$=30). \textbf{Bold} = best overall; \underline{underlined} = second best.}
\label{tab:oi_results}
\begin{tabular}{@{}ll|cccc|cc|cc|c@{}}
\toprule
& & \multicolumn{4}{c|}{\textbf{Simple} ($N$=700)} & \multicolumn{2}{c|}{\textbf{Intermediate} ($N$=200)} & \multicolumn{2}{c|}{\textbf{Hard} ($N$=110)} & \\
\cmidrule(lr){3-6} \cmidrule(lr){7-8} \cmidrule(lr){9-10}
\textbf{Model} & \textbf{Reasoning} & OC & OPD & OIn &  OPl & RATA & OS & ORA & SIT & \textbf{Overall} \\
\midrule
\multicolumn{11}{c}{\cellcolor[HTML]{FFF3CA}\textbf{Closed-Source}} \\
\midrule
GPT-5 & high & 89.7 & 73.7 & 66.3 & 71.4 & 36.4 & \underline{34.0} & 40.8 & \underline{76.7} & 61.1 \\
GPT-5 & low & 90.3 & 72.0 & 70.9 & 71.4 & 30.8 & 28.0 & 40.4 & 73.3 & 59.6 \\
GPT-5 Pro & high & \textbf{92.0} & 73.1 & 71.4 & 70.9 & 36.1 & 29.0 & 42.5 & \textbf{80.0} & 61.9 \\
GPT-5.2 Pro & high & 88.6 & 76.0 & \underline{72.6} & 72.0 & 34.6 & 28.0 & \textbf{52.8} & 73.3 & 62.2 \\
GPT-OSS-120B & high & 82.9 & 60.6 & 65.1 & 72.0 & 25.1 & \underline{34.0} & 35.6 & 56.7 & 54.0 \\
o3 & high & 89.1 & 70.9 & 68.0 & 70.3 & 31.5 & 32.0 & 42.4 & 70.0 & 59.3 \\
o4-mini & high & 88.6 & 65.7 & 69.1 & 73.7 & 29.0 & 32.0 & 40.5 & 73.3 & 59.0 \\
\midrule
Gemini 2.5 Pro & 16K & 89.7 & \underline{78.9} & 68.0 & 72.6 & 30.3 & 31.0 & 39.7 & 66.7 & 59.6 \\
Gemini 2.5 Pro & 32K & 87.4 & 78.3 & 66.9 & 72.0 & 34.0 & 27.0 & 42.4 & 70.0 & 59.7 \\
Gemini 2.5 Pro & 8K & 88.6 & \textbf{80.0} & 65.7 & 73.7 & 31.5 & 29.0 & 37.9 & 63.3 & 58.7 \\
\midrule
Grok 3 Mini & high & 81.7 & 72.0 & 68.0 & 73.7 & 37.0 & 22.0 & 41.9 & 66.7 & 57.9 \\
Grok 3 Mini & low & 81.7 & 73.1 & 66.3 & 72.6 & 36.0 & 18.0 & 37.2 & 73.3 & 57.3 \\
Grok 4.1 Fast & default & 88.6 & 67.4 & 66.9 & 70.3 & 35.5 & 33.0 & 31.1 & 73.3 & 58.3 \\
\midrule
Claude Opus 4.5 & high & \textbf{92.0} & 76.6 & 71.4 & 73.1 & \textbf{42.2} & 25.0 & 45.8 & 70.0 & 62.0 \\
Claude Opus 4.6 & high & \underline{91.4} & 78.3 & 71.4 & \textbf{74.9} & \underline{40.0} & 26.0 & 49.6 & 73.3 & \underline{63.1} \\
Claude Opus 4.6 & max & \textbf{92.0} & 77.7 & \textbf{74.9} & \underline{74.3} & 38.9 & 26.0 & \underline{51.1} & \textbf{80.0} & \textbf{64.4} \\
Claude Sonnet 4.5 & — & 89.1 & 67.4 & 66.9 & 71.4 & 34.9 & 29.0 & 38.4 & \textbf{80.0} & 59.6 \\
\midrule
\multicolumn{11}{c}{\cellcolor[HTML]{D9E1F4}\textbf{Open-Source}} \\
\midrule
DeepSeek Reasoner & 16K & 79.4 & 69.7 & 68.6 & \textbf{74.9} & 36.0 & 25.0 & 29.1 & 70.0 & 56.6 \\
DeepSeek Reasoner & 32K & 81.1 & 73.7 & 69.7 & 73.1 & 33.1 & 32.0 & 31.6 & 73.3 & 58.5 \\
DeepSeek Reasoner & 8K & 80.6 & 70.3 & 71.4 & 72.6 & 34.7 & \textbf{35.0} & 30.5 & 63.3 & 57.3 \\
Llama 3.3 70B & — & 83.4 & 60.6 & 68.0 & 72.0 & 26.8 & 29.0 & 35.0 & 46.7 & 52.7 \\
\bottomrule
\end{tabular}
\end{table*}

\begin{figure}[!hb]
  \centering
  \includegraphics[width=\linewidth]{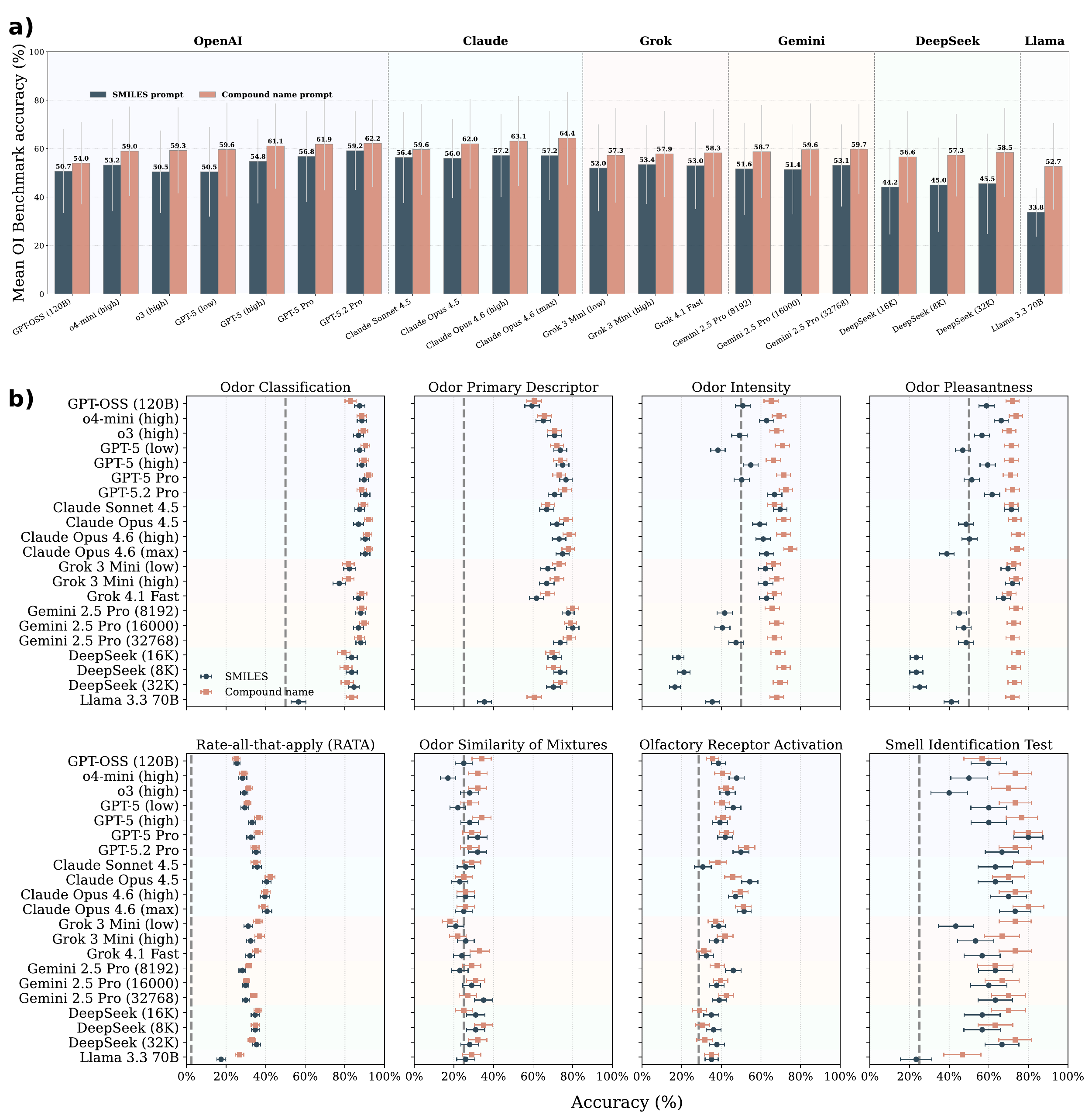}
  \caption{\textbf{Olfactory Perception (OP) benchmark performance} \textbf{(a)}
Two bars per model show overall mean accuracy across all categories for isomeric SMILES and compound name prompts. Multi‑answer categories use multilabel F1; all others use any‑overlap accuracy. Error bars indicate 95\% confidence intervals of the cross‑category mean. Models are grouped by LLM family with shaded bands and dashed boundaries; labels use shortened model names. \textbf{(b)} Eight subplots show per‑category accuracy for each model, with separate isomeric SMILES and compound‑name prompts. Horizontal error bars are bootstrap standard deviations of the mean accuracy. Each panel includes a category‑specific chance baseline shown by gray dashed line.}
  \label{fig:figure2}
\end{figure}


\section{Results}

In this section, we evaluate a diverse set of state-of-the-art LLMs on the OP benchmark to quantify current olfactory reasoning ability and identify systematic failure modes. We report results across all task categories and compare prompting conditions (compound names vs. isomeric \ SMILES) to isolate the effect of molecular representation on performance.

\subsection{Overall Performance}

Table~\ref{tab:oi_results} and Figure~\ref{fig:figure2} present benchmark performance across all evaluated models. The best-performing configuration, \textbf{Claude Opus 4.6 (max)}, attains 64.4\% overall accuracy
with compound name prompts, followed by \textbf{Claude Opus 4.6 (high)} at 63.1\%, \textbf{GPT-5.2 Pro} at 62.2\%, and \textbf{Claude Opus 4.5} at 62.0\%. These scores substantially exceed chance levels for each task category (Figure~\ref{fig:figure2}b, dashed lines), yet remain far from ceiling performance, indicating that, while frontier models encode meaningful olfactory knowledge, substantial room for improvement remains.

A clear performance hierarchy emerges across providers. Anthropic and OpenAI frontier models occupy the top positions, with all configurations except GPT-OSS-120B surpassing 59\% accuracy. The progression within the Claude family, from Sonnet 4.5 (59.6\%) through Opus 4.5 (62.0\%) to Opus 4.6 (max) (64.4\%), reflects consistent improvements associated with increased model capability and reasoning depth. OpenAI's GPT-5 variants demonstrate competitive performance, with GPT-5.2 Pro trailing Claude Opus 4.6 (max) by 2.2 percentage points.
The o-series reasoning models perform well, with o3 (high) at 59.3\% and o4-mini (high) at 59.0\%. Google's Gemini 2.5 Pro (58.7--59.7\%)
and xAI's Grok family (57.3--58.3\%) occupy the mid-tier, while DeepSeek Reasoner reaches 56.6--58.5\%.
The open-source Llama 3.3 70B lags behind all proprietary alternatives at 52.7\%,
underscoring a persistent capability gap between open-weight and closed-source systems for specialized perceptual reasoning.

Performance varies considerably across task categories. Simple tasks (OC, OPD, OIn, OPl) produce the highest accuracy, with top models reaching 92.0\% on odor classification and 80.0\% on primary descriptor identification. Intermediate tasks (RATA, OS) prove more demanding, with best performance limited to 42.2\% and 35.0\% respectively. Hard tasks exhibit the widest variance: SIT reaches 80.0\% for three models, benefiting from knowledge about foods and beverages, whereas ORA peaks at 52.8\% (GPT-5.2 Pro),
still a challenging task due to the specialized biochemistry knowledge required. A detailed question-level difficulty analysis is provided in 
Appendix~\ref{appendix:difficulty_distribution} 
(Figure~\ref{fig:difficulty_distribution}) for single-label tasks 
and Appendix~\ref{appendix:multilabel_difficulty} 
(Figure~\ref{fig:rata_ora_difficulty}) for the multi-label tasks 
RATA and ORA.

\subsection{Isomeric SMILES vs.\ Compound Name Prompts}

A consistent and substantial performance gap separates the two molecular representation formats (Figure~\ref{fig:figure2}a). Across all 21 model configurations, compound name prompts outperform isomeric SMILES notation, with improvements ranging from 2.4 percentage points (Claude Opus 4.5: 59.6\% $\rightarrow$ 62.0\%) to 18.9 percentage points (Llama 3.3 70B: 33.8\% $\rightarrow$ 52.7\%).
DeepSeek Reasoner (8K) shows a 12.3-point improvement (45.0\% $\rightarrow$ 57.3\%).
The mean improvement across models is approximately 7 percentage points, suggesting that current LLMs access olfactory knowledge primarily through lexical associations rather than structural molecular reasoning.

This representation gap exhibits systematic patterns linked to model capability. Frontier reasoning models display the narrowest gaps: GPT-5 Pro, GPT-5.2 Pro, and Claude Opus 4.5 maintain strong isomeric SMILES performance, each reaching over 95\% of their compound name accuracy. Claude Opus 4.6 (max) retains over 92\% of its name-based score despite achieving the highest overall accuracy. Conversely, Llama 3.3 70B exhibits the largest disparity, with isomeric SMILES accuracy barely exceeding chance while compound name performance reaches 52.7\%.
DeepSeek models present an intermediate case: despite limited isomeric SMILES performance (44-46\%),
they attain competitive compound name accuracy (57-59\%),
which indicates robust lexical knowledge but constrained structural reasoning capabilities.

Per-task analysis (Figure~\ref{fig:figure2}b) reveals that the isomeric SMILES/name gap varies substantially by task category.  OC exhibits the smallest gaps, potentially because odorousness correlates with molecular properties (volatility,
molecular weight) that can be partially inferred from isomeric SMILES structure. OIn, OPl, and SIT show the largest gaps, consistent with tasks where identifying the target molecule or food source is critical and far easier from a name than from structural notation.  In contrast, multi-label tasks (RATA, ORA) show modest gaps, suggesting that descriptor-level knowledge is similarly accessible, or similarly limited, under both representations.

\subsection{Effect of Reasoning Budget}

Systematic variation of reasoning token budgets reveals consistent but modest performance gains. Within the GPT-5 family, increasing from low to high reasoning improves compound name accuracy from 59.6\% to 61.1\% (+1.5 points), while GPT-5 Pro reaches 61.9\% (+0.8 points additional), and GPT-5.2 Pro attains 62.2\%. Claude Opus 4.6 shows consistent scaling: the high configuration (63.1\%) improves over Claude Opus 4.5 (62.0\%), and the max setting reaches 64.4\%.

Similar patterns appear across other providers. Gemini 2.5 Pro improves from 58.7\% (8K tokens) to 59.6\% (16K tokens) to 59.7\% (32K tokens).
Grok 3 Mini advances from 57.3\% (low) to 57.9\% (high). DeepSeek Reasoner shows a non-monotonic pattern: 57.3\% at 8K, 56.6\% at 16K, and 58.5\% at 32K,
suggesting that optimal reasoning budgets may be task-dependent.

Overall, no model gains more than approximately 2 percentage points from extended reasoning on this benchmark. This contrasts with findings in chemistry benchmarks where reasoning continued to provide substantial gains at higher budgets~\cite{runcie2025assessing}. This may reflect the more constrained nature of olfactory knowledge compared to general chemical reasoning.
\begin{figure}[!hb]
  \centering
  \includegraphics[width=0.86\linewidth]{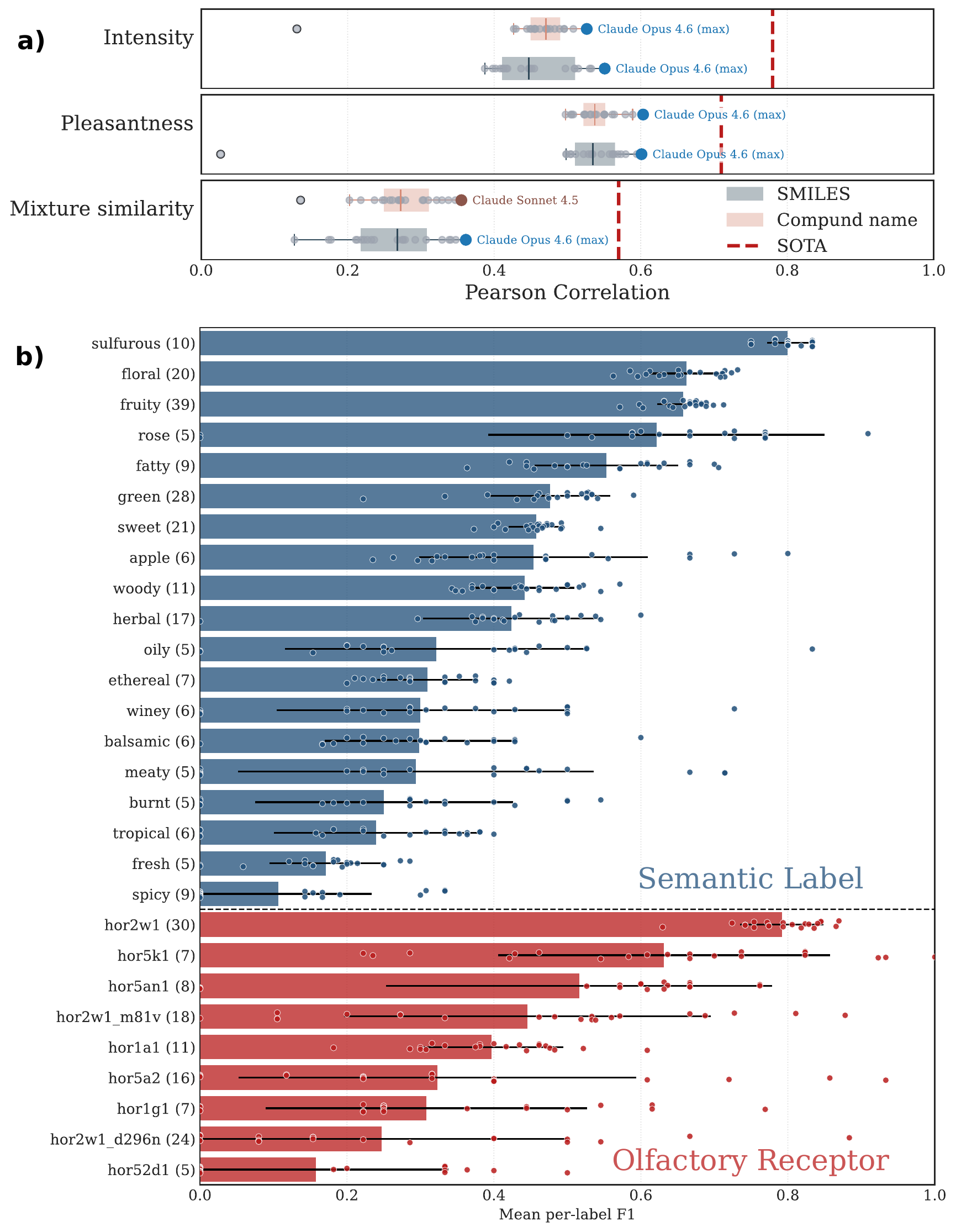}
  \caption{\textbf{Predictions correlations and performance difficulty across olfactory tasks.} \textbf{(a)} Pearson correlations across models for three continuous‑rating categories (odor intensity, odor pleasantness, and mixture similarity). Each row shows the distribution of model correlations for isomeric SMILES and Compound name prompts; gray points represent individual models, and the best‑performing model per prompt/category is highlighted with a colored circle. State‑of‑the‑art reference performance is indicated by red dashed lines. \textbf{(b)} Combined label‑difficulty ranking for RATA (blue) and olfactory‑receptor activation (red). For each label, per‑model F1 scores are computed in a multilabel setting; the bar shows the mean F1 across models with an error bar (standard deviation), while  points represent individual model values.}
  \label{fig:figure4}
\end{figure}

\subsection{Fine-Grained Performance Analysis}
\label{sec:finegrained}
Beyond categorical accuracy, we assess whether models capture the 
continuous structure of human olfactory perception.  
Figure~\ref{fig:figure4}a presents Pearson correlations between 
model-predicted ratings and human psychophysical measurements: the best 
models reach $r \approx 0.55$ for OIn, approaching specialized model 
performance \cite{keller2017predicting,satarifard2025high} (red dashed line);  OPl correlations are higher 
($r \approx 0.60$), 
and OS exhibits the weakest correlations 
($r \approx 0.35$), confirming that models struggle to integrate 
perceptual information across molecules. Isomeric SMILES and compound name prompts produce 
similar correlations across all three dimensions, with a slightly larger gap for OIn.  

Turning to the multi-label tasks, RATA and ORA 
exhibit distinct difficulty profiles: RATA F1 scores are mainly bell-shaped, 
indicating partial credit on most questions, while ORA is 
bimodal: models either possess the relevant receptor-ligand knowledge 
or lack it entirely (Appendix~\ref{appendix:multilabel_difficulty}, 
Figure~\ref{fig:rata_ora_difficulty}). Per-label analysis 
(Figure~\ref{fig:figure4}b) reveals a clear divide for RATA: 
descriptors with well-established functional-group 
associations (e.g. sulfurous, floral, fruity \cite{genva2019possible}) achieve the highest mean F1, while descriptors such as spicy, fresh, tropical prove nearly impossible. A 
detailed case study of this failure mode is provided in 
Appendix~\ref{appendix:per_label_difficulty}. For ORA, per-label 
difficulty varies sharply across receptors 
(Figure~\ref{fig:figure4}b): the wildtype \texttt{hOR2W1} (appearing in 30 label assignments) is reliably predicted at F1 above 0.6, and the 
\texttt{M81V} variant is moderately well-predicted, but 
\texttt{hOR2W1\_D296N} (24) and \texttt{hOR52D1} (5) sit near zero for most models. 
The \texttt{hOR52D1} gap 
likely reflects data scarcity, while \texttt{D296N} reveals a specific 
knowledge error: Claude Opus~4.6 (max) never predicts this receptor 
across all 24 label appearances, whereas GPT-5.2~Pro correctly identifies it 
in 19 cases. Inter-model variance is substantially higher for receptor 
labels than for semantic descriptors, indicating that receptor 
knowledge is more idiosyncratic across model families; a detailed 
analysis of these knowledge gaps is provided in 
Appendix~\ref{appendix:ora_failure} 
(Figure~\ref{fig:case_study_D296N}).

Two qualitatively distinct failure mechanisms underlie the hardest 
categories.  For OS, models use molecular overlap as a proxy for 
perceptual similarity: accuracy reaches $\sim$85\% when similar 
mixtures share many molecules but drops to near 0\% when similar 
mixtures share few, and all models exhibit a systematic bias toward 
predicting dissimilarity, Claude models assign ``Slightly Dissimilar'' 
to the vast majority of mixtures, even though perceptually similar 
mixtures often share zero molecules.  This heuristic renders 
mixture-level olfactory similarity fundamentally beyond current LLM 
capabilities; a detailed analysis is provided in 
Appendix~\ref{appendix:OS_subcategory_accuracy} 
(Figures~\ref{fig:os_scatter},~\ref{fig:os_confusion},~\ref{fig:os_case_study}).  
Additionally, Claude Opus~4.6's safety filter refuses OC questions about hazardous compounds (e.g., nerve agents), illustrating a tension between safety alignment and scientific evaluation (Appendix~\ref{appendix:safety_alignment}).

\begin{figure}[t]
  \centering
  \includegraphics[width=1.0\linewidth]{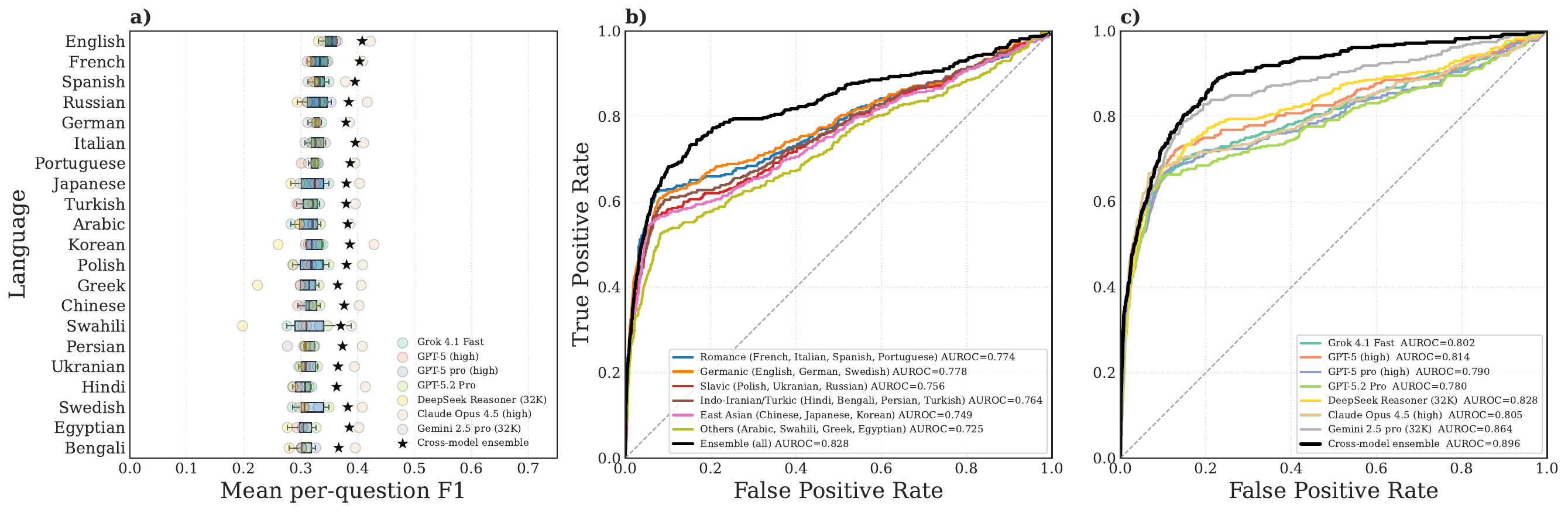}
  \caption{\textbf{Multilingual RATA performance for compound name prompt.} \textbf{(a)} Per-language distributions of model performance, summarized as mean per-question multilabel F1, colored points denote individual models and the black star denotes a cross-model ensemble. \textbf{(b)} DeepSeek (32K) AUROC by language family using per-label vote fractions across languages within each family (East Asian ans Others are not same language families), black curve represent an ensemble of all languages. \textbf{(c)} AUROC for each model using an all-language majority vote ensemble, plus a cross-model ensemble pooling all models from all languages. Dashed diagonal indicates chance.}
  \label{fig:figure5}
\end{figure}

\subsection{Multilingual Evaluation}
\label{sec:multilingualeval}

To probe whether olfactory knowledge is language-specific, we translated
the RATA task prompts into 21 languages spanning over six language families
and evaluated seven models per language.  Figure~\ref{fig:figure5}a
shows that English achieves the highest mean per-question F1, followed
closely by French, Spanish, and Russian, while
non-Indo-European languages (Korean, Chinese, Swahili) cluster
at the lower end.

Figure~\ref{fig:figure5}b presents language-family AUROC curves for
vote-fraction of DeepSeek (32K) model: Germanic languages (English, German, Swedish) achieve the
highest family-level AUROC (0.778), followed closely by Romance (French,
Italian, Spanish, Portuguese; 0.774), Indo-Iranian/Turkic (Hindi,
Bengali, Persian, Turkish; 0.764), Slavic (Polish, Ukrainian, Russian;
0.756), East Asian (Chinese, Japanese, Korean; 0.749), and the
``Others'' group (Arabic, Swahili, Greek, Egyptian; 0.725).  An ensemble
aggregating votes across all 21 languages achieves AUROC\,=\,0.828,
suggesting that multilingual aggregation captures complementary olfactory
knowledge.

Figure~\ref{fig:figure5}c shows per-model AUROC using an all-language
vote-fraction ensemble: Gemini 2.5~Pro (32K) leads at 0.864, followed
by DeepSeek Reasoner (32K) at 0.828, GPT-5 (high) at 0.814,
Claude Opus~4.5 (high) at 0.805, Grok~4.1 Fast at 0.802,
GPT-5~Pro (high) at 0.790, and GPT-5.2~Pro at 0.780.  A cross-model
ensemble combining all seven models reaches 0.896, indicating that model
diversity further improves multilingual olfactory prediction beyond
language diversity alone.

\section{Conclusion, Limitations, and Future Work}
Here, we introduce the Olfactory Perception (OP) benchmark, a structured evaluation suite for testing whether large language models can reason about smell from molecular information and real-world odor sources. Across a broad set of olfactory tasks, we observe emerging but incomplete capabilities: the best-performing systems achieve moderate accuracy, yet performance remains far from reliable and varies substantially by task type. A consistent and striking pattern is the gap between prompts using common compound names and prompts using isomeric SMILES, which suggests current LLMs often succeed via lexical associations rather than robust reasoning over molecular structure. Taken together, our results position olfaction as a challenging and underexplored modality for LLM evaluation, and provide a concrete benchmark for measuring progress toward models that can connect chemistry to sensory perception.

The OP benchmark is intended as a controlled, scalable probe of olfactory knowledge and reasoning in LLMs, and it makes a set of deliberate design choices. We focus on standardized descriptor vocabularies and curated judgments, which enables consistent automatic evaluation but does not capture the full richness of odor perception (e.g., contextual effects, individual and cultural variation). Our tasks use discrete response formats (e.g., fixed descriptor sets), prioritizing reproducibility and comparability across models, while potentially under-weighting nuanced or partially correct free-form descriptions. As with other instruction-following benchmarks, results can be influenced by prompting and output formatting sensitivity, despite using consistent templates. Multilingual variants are produced via automated translation with validation, which improves coverage but may introduce subtle shifts in connotation and usage frequency across languages. Finally, benchmark accuracy should be interpreted as agreement with established labels rather than as evidence of mechanistic olfactory understanding or experimentally verified predictions.

Several directions could strengthen both the benchmark and model evaluation. On the dataset side, expanding coverage to more mixtures, concentration effects, temporal dynamics, and broader receptor-level data would better reflect real olfaction. Incorporating human evaluation and cross-cultural/individual annotation (including synonym sets and graded scoring) could capture partial correctness and reduce brittleness to surface forms. Incorporating multimodal inputs, such as audio 
recordings of verbal odor descriptions and images of facial expressions 
during smelling, could further bridge the gap between molecular 
representations and embodied olfactory experience. On the modeling side, it will be valuable to study hybrid systems that combine LLMs with cheminformatics tools and learned molecular representations, and to develop methods that explicitly encourage structure-based reasoning (e.g., controlling for name memorization, adversarial splits, and representation-invariant prompting).  Prompt engineering strategies such 
as prompt repetition~\cite{leviathan2025prompt}, which improves 
non-reasoning LLM performance without additional generation cost, could 
also be explored as a low-cost approach to improving olfactory task 
accuracy. Finally, extending evaluation to generative settings (e.g., proposing molecules with target odor profiles under safety and synthesizability constraints) would connect benchmark performance to practical applications in fragrance, flavor, and environmental monitoring.



\FloatBarrier

\bibliographystyle{unsrt}  
\bibliography{references}

\clearpage
\appendix

\appendix

\section{Prompt Templates}
\label{appendix:prompts}

This appendix presents the exact prompt templates used for each question category in the OP benchmark, together with one real example drawn directly from the benchmark dataset. Each task is evaluated using two prompt variants: one using isomeric SMILES molecular representations and one using common compound names. All prompts instruct models to respond without additional commentary to facilitate automated answer extraction.

\subsection{Odor Classification}
\label{appendix:odor_classification}

The odor classification task tests whether models can determine if a molecule has a detectable odor to humans ($n=175$).

\begin{tcolorbox}[
    colback=blue!5!white,
    colframe=blue!75!black,
    title=Template -- Isomeric SMILES Prompt,
    fonttitle=\bfseries,
    breakable
]
\texttt{Does [\{SMILES\}] have a detectable odor to humans? Only respond with Odorous or Odorless. Do not write any comments.}
\end{tcolorbox}

\begin{tcolorbox}[
    colback=green!5!white,
    colframe=green!75!black,
    title=Template -- Name Prompt,
    fonttitle=\bfseries,
    breakable
]
\texttt{Does [\{compound\_name\}] have a detectable odor to humans? Only respond Odorous or Odorless. Do not write any comments.}
\end{tcolorbox}

\begin{tcolorbox}[
    colback=gray!5!white,
    colframe=gray!70!black,
    title=Example: Phenacetin (Odorless),
    fonttitle=\bfseries,
    breakable
]
\small
\textbf{Isomeric SMILES prompt:}\\
\texttt{Does [CCOC1=CC=C(C=C1)NC(=O)C] have a detectable odor to humans? Only respond with Odorous or Odorless. Do not write any comments.}

\medskip
\textbf{Name prompt:}\\
\texttt{Does [Phenacetin] have a detectable odor to humans? Only respond Odorous or Odorless. Do not write any comments.}

\medskip
\textbf{Ground-truth answer:} Odorless

\medskip
\textbf{Model predictions (compound name prompt):}\\
\begin{tabular}{lll}
Gemini 2.5 Pro & Odorless & \textcolor{green!60!black}{\checkmark} \\
GPT-5 Pro & Odorless & \textcolor{green!60!black}{\checkmark} \\
o3 (high) & Odorless & \textcolor{green!60!black}{\checkmark} \\
\end{tabular}
\end{tcolorbox}

\subsection{Odor Primary Descriptor}
\label{appendix:primary_descriptor}

This task requires models to identify the single odor primary descriptor for a molecule from four options ($n=175$). The four options consist of the correct answer and three distractor descriptors.

\begin{tcolorbox}[
    colback=blue!5!white,
    colframe=blue!75!black,
    title=Template -- Isomeric SMILES Prompt,
    fonttitle=\bfseries,
    breakable
]
\texttt{What is the odor primary descriptor of molecule [\{SMILES\}]? Select only one from the [\{OPTIONS\}]. Do not write any comments.}

\vspace{0.5em}
\end{tcolorbox}

\begin{tcolorbox}[
    colback=green!5!white,
    colframe=green!75!black,
    title=Template -- Name Prompt,
    fonttitle=\bfseries,
    breakable
]
\texttt{What is the odor primary descriptor of molecule [\{compound\_name\}]? Select only one from the [\{OPTIONS\}]. Do not write any comments.}
\end{tcolorbox}

\begin{tcolorbox}[
    colback=gray!5!white,
    colframe=gray!70!black,
    title=Example: Octanal diethyl acetal,
    fonttitle=\bfseries,
    breakable
]
\small
\textbf{Isomeric SMILES prompt:}\\
\texttt{What is the odor primary descriptor of molecule [CCCCCCCC(OCC)OCC]? Select only one from the [Green;Honey;Herbal;Marine]. Do not write any comments.}

\medskip
\textbf{Name prompt:}\\
\texttt{What is the odor primary descriptor of molecule [Octanal diethyl acetal]? Select only one from the [Green;Honey;Herbal;Marine]. Do not write any comments.}

\medskip
\textbf{Ground-truth answer:} Green

\medskip
\textbf{Model predictions (compound name prompt):}\\
\begin{tabular}{lll}
Gemini 2.5 Pro & Green & \textcolor{green!60!black}{\checkmark} \\
GPT-5 Pro & Green & \textcolor{green!60!black}{\checkmark} \\
o3 (high) & Green & \textcolor{green!60!black}{\checkmark} \\
\end{tabular}
\end{tcolorbox}

\subsection{Odor Intensity}
\label{appendix:odor_intensity}

This task presents pairs of molecules and asks models to identify which has higher perceived odor intensity, plus provide numerical intensity estimates ($n=175$).

\begin{tcolorbox}[
    colback=blue!5!white,
    colframe=blue!75!black,
    title=Template -- Isomeric SMILES Prompt,
    fonttitle=\bfseries,
    breakable
]
\texttt{Which molecule is likely to smell more intense to humans: \{SMILES\_1\} or \{SMILES\_2\}? Select only one of the SMILES. If you had to rate the intensity of these molecules from 0 (extremely low) to 100 (highly intense), what would you assign to each? Respond with the selected compound name (the one with higher intensity), followed by two intensity values in the order the molecules are listed. Use semicolons (;) as separators. Do not write any comments.}
\end{tcolorbox}

\begin{tcolorbox}[
    colback=green!5!white,
    colframe=green!75!black,
    title=Template -- Name Prompt,
    fonttitle=\bfseries,
    breakable
]
\texttt{Which molecule is likely to smell more intense to humans: \{compound\_name\_1\} or \{compound\_name\_2\}? Select only one of the compound names. If you had to rate the intensity of these molecules from 0 (extremely low) to 100 (highly intense), what would you assign to each? Respond with the selected compound name (the one with higher intensity), followed by two intensity values in the order the molecules are listed. Use semicolons (;) as separators. Do not write any comments.}
\end{tcolorbox}

\begin{tcolorbox}[
    colback=gray!5!white,
    colframe=gray!70!black,
    title=Example: hexan-2-one vs.\ pyrazine,
    fonttitle=\bfseries,
    breakable
]
\small
\textbf{Isomeric SMILES prompt:}\\
\texttt{Which molecule is likely to smell more intense to humans: CCCCC(=O)C or C1=CN=CC=N1? Select only one of the SMILES. If you had to rate the intensity of these molecules from 0 (extremely low) to 100 (highly intense), what would you assign to each? Respond with the selected compound name (the one with higher intensity), followed by two intensity values in the order the molecules are listed. Use semicolons (;) as separators. Do not write any comments.}

\medskip
\textbf{Name prompt:}\\
\texttt{Which molecule is likely to smell more intense to humans: hexan-2-one or pyrazine? Select only one of the compound names. If you had to rate the intensity of these molecules from 0 (extremely low) to 100 (highly intense), what would you assign to each? Respond with the selected compound name (the one with higher intensity), followed by two intensity values in the order the molecules are listed. Use semicolons (;) as separators. Do not write any comments.}

\medskip
\textbf{Ground-truth answer:} hexan-2-one (intensity = 72.4 vs.\ 20.8)\\
\textbf{Expected response format:} \texttt{hexan-2-one;72;21}

\medskip
\textbf{Model predictions (compound name prompt):}\\
\begin{tabular}{lll}
Gemini 2.5 Pro & \texttt{Pyrazine; 45; 95} & \textcolor{red!70!black}{\ding{55}} \\
GPT-5 Pro & \texttt{pyrazine;30;70} & \textcolor{red!70!black}{\ding{55}} \\
o3 (high) & \texttt{pyrazine;30;85} & \textcolor{red!70!black}{\ding{55}} \\
\end{tabular}\\
\textit{All three models incorrectly select pyrazine as more intense.}
\end{tcolorbox}

\subsection{Odor Pleasantness}
\label{appendix:odor_pleasantness}

This task presents pairs of molecules and asks models to identify which smells more pleasant, plus provide numerical pleasantness estimates ($n=175$).

\begin{tcolorbox}[
    colback=blue!5!white,
    colframe=blue!75!black,
    title=Template -- Isomeric SMILES Prompt,
    fonttitle=\bfseries,
    breakable
]
\texttt{Which molecule is likely to smell more pleasant to humans: \{SMILES\_1\} or \{SMILES\_2\}? Select only one of the SMILES. If you had to rate the pleasantness of these molecules from 0 (extremely unpleasant) to 100 (highly pleasant), what would you assign to each? Respond with the selected compound name (the one with higher pleasantness), followed by two pleasantness values in the order the molecules are listed. Use semicolons (;) as separators. Do not write any comments.}
\end{tcolorbox}

\begin{tcolorbox}[
    colback=green!5!white,
    colframe=green!75!black,
    title=Template -- Name Prompt,
    fonttitle=\bfseries,
    breakable
]
\texttt{Which molecule is likely to smell more pleasant to humans: \{compound\_name\_1\} or \{compound\_name\_2\}? Select only one of the compound names. If you had to rate the pleasantness of these molecules from 0 (extremely unpleasant) to 100 (highly pleasant), what would you assign to each? Respond with the selected compound name (the one with higher pleasantness), followed by two pleasantness values in the order the molecules are listed. Use semicolons (;) as separators. Do not write any comments.}
\end{tcolorbox}

\begin{tcolorbox}[
    colback=gray!5!white,
    colframe=gray!70!black,
    title=Example: 2-hydroxybenzaldehyde vs.\ ethyl hexanoate,
    fonttitle=\bfseries,
    breakable
]
\small
\textbf{Isomeric SMILES prompt:}\\
\texttt{Which molecule is likely to smell more pleasant to humans: C1=CC=C(C(=C1)C=O)O or CCCCCC(=O)OCC? Select only one of the SMILES. If you had to rate the pleasantness of these molecules from 0 (extremely unpleasant) to 100 (highly pleasant), what would you assign to each? Respond with the selected compound name (the one with higher pleasantness), followed by two pleasantness values in the order the molecules are listed. Use semicolons (;) as separators. Do not write any comments.}

\medskip
\textbf{Name prompt:}\\
\texttt{Which molecule is likely to smell more pleasant to humans: 2-hydroxybenzaldehyde or ethyl hexanoate? Select only one of the compound names. If you had to rate the pleasantness of these molecules from 0 (extremely unpleasant) to 100 (highly pleasant), what would you assign to each? Respond with the selected compound name (the one with higher pleasantness), followed by two pleasantness values in the order the molecules are listed. Use semicolons (;) as separators. Do not write any comments.}

\medskip
\textbf{Ground-truth answer:} ethyl hexanoate (pleasantness = 79.2 vs.\ 29.7)\\
\textbf{Expected response format:} \texttt{ethyl hexanoate;30;79}

\medskip
\textbf{Model predictions (compound name prompt):}\\
\begin{tabular}{lll}
Gemini 2.5 Pro & \texttt{Ethyl hexanoate; 65; 90} & \textcolor{green!60!black}{\checkmark} \\
GPT-5 Pro & \texttt{ethyl hexanoate;55;90} & \textcolor{green!60!black}{\checkmark} \\
o3 (high) & \texttt{ethyl hexanoate;35;85} & \textcolor{green!60!black}{\checkmark} \\
\end{tabular}\\
\textit{All models correctly select ethyl hexanoate, though numerical estimates vary.}
\end{tcolorbox}

\subsection{Rate-All-That-Apply (RATA)}
\label{appendix:rata}

This multi-label classification task requires models to select all applicable odor descriptors from a comprehensive list of 138 possible descriptors ($n=100$).

\begin{tcolorbox}[
    colback=blue!5!white,
    colframe=blue!75!black,
    title=Template -- Isomeric SMILES Prompt,
    fonttitle=\bfseries,
    breakable
]
\texttt{Which of the following odor descriptors apply to molecule \{SMILES\}? Select all that apply from [\{138 descriptors\}]. Do not write any comments.}

\vspace{0.5em}
\end{tcolorbox}

\begin{tcolorbox}[
    colback=green!5!white,
    colframe=green!75!black,
    title=Template -- Name Prompt,
    fonttitle=\bfseries,
    breakable
]
\texttt{Which of the following odor descriptors apply to molecule \{compound\_name\}? Select all that apply from [\{138 descriptors\}]. Do not write any comments.}
\end{tcolorbox}

\begin{tcolorbox}[
    colback=gray!5!white,
    colframe=gray!70!black,
    title=Example: 2{,}3-Dimethylpentanal,
    fonttitle=\bfseries,
    breakable
]
\small
\textbf{Isomeric SMILES prompt:}\\
\texttt{Which of the following odor descriptors apply to molecule CCC(C)C(C)C=O? Select all that apply from [alcoholic; aldehydic; alliaceous; almond; amber; animal; anisic; apple; apricot; aromatic; balsamic; banana; beefy; bergamot; berry; bitter; black currant; brandy; burnt; buttery; cabbage; camphoreous; caramellic; cedar; celery; chamomile; cheesy; cherry; chocolate; cinnamon; citrus; clean; clove; cocoa; coconut; coffee; cognac; cooked; cooling; cortex; coumarinic; creamy; cucumber; dairy; dry; earthy; ethereal; fatty; fermented; fishy; floral; fresh; fruit skin; fruity; garlic; gassy; geranium; grape; grapefruit; grassy; green; hawthorn; hay; hazelnut; herbal; honey; hyacinth; jasmin; juicy; ketonic; lactonic; lavender; leafy; leathery; lemon; lily; malty; meaty; medicinal; melon; metallic; milky; mint; muguet; mushroom; musk; musty; natural; nutty; odorless; oily; onion; orange; orangeflower; orris; ozone; peach; pear; phenolic; pine; pineapple; plum; popcorn; potato; powdery; pungent; radish; raspberry; ripe; roasted; rose; rummy; sandalwood; savory; sharp; smoky; soapy; solvent; sour; spicy; strawberry; sulfurous; sweaty; sweet; tea; terpenic; tobacco; tomato; tropical; vanilla; vegetable; vetiver; violet; warm; waxy; weedy; winey; woody]. Do not write any comments.}

\medskip
\textbf{Name prompt:}\\
\texttt{Which of the following odor descriptors apply to molecule 2,3-Dimethylpentanal? Select all that apply from [\ldots same 138 descriptors \ldots]. Do not write any comments.}

\medskip
\textbf{Ground-truth answer:} ethereal; green\\
\textbf{Expected response format:} \texttt{ethereal;green}

\medskip
\textbf{Model predictions (compound name prompt):}\\
\begin{tabular}{p{2.5cm}p{8.5cm}l}
Gemini 2.5 Pro & \texttt{aldehydic; chocolate; citrus; cocoa; cortex; fatty; green; malty; oily; woody} & \textcolor{orange!80!black}{$\sim$} \\
GPT-5 Pro & \texttt{aldehydic; citrus; fresh; fruity; green; fatty; oily; apple} & \textcolor{orange!80!black}{$\sim$} \\
o3 (high) & \texttt{aldehydic; burnt; chocolate; cocoa; malty; nutty; roasted} & \textcolor{red!70!black}{\ding{55}} \\
\end{tabular}\\
\textit{Gemini and GPT-5 Pro partially overlap with ground truth (green), but also predict many false positives. o3 misses both ground-truth descriptors entirely.}
\end{tcolorbox}

\subsection{Odor Similarity of Mixtures}
\label{appendix:mixture_similarity}

This task evaluates models' ability to judge perceptual similarity between two odor mixtures containing 2--10 molecules each ($n=100$).

\begin{tcolorbox}[
    colback=blue!5!white,
    colframe=blue!75!black,
    title=Template -- Isomeric SMILES Prompt,
    fonttitle=\bfseries,
    breakable
]
\texttt{Mixture A contains molecules [\{SMILES\_list\_A\}] and mixture B contains [\{SMILES\_list\_B\}]. On the scale [Strongly Similar; Slightly Similar; Slightly Dissimilar; Strongly Dissimilar], select how similar these mixtures smell. If you had to rate the olfactory perceptual distance on a 0.0--1.0 scale (0.0 = identical, 1.0 = completely different), what distance do you assign? Respond with your selection from [Strongly Similar; Slightly Similar; Slightly Dissimilar; Strongly Dissimilar], followed by the distance value with two-decimal precision. Use semicolons (;) as separators. Do not write any comments.}
\end{tcolorbox}

\begin{tcolorbox}[
    colback=green!5!white,
    colframe=green!75!black,
    title=Template -- Name Prompt,
    fonttitle=\bfseries,
    breakable
]
\texttt{Mixture A contains molecules [\{name\_list\_A\}] and mixture B contains [\{name\_list\_B\}]. On the scale [Strongly Similar; Slightly Similar; Slightly Dissimilar; Strongly Dissimilar], select how similar these mixtures smell. If you had to rate the olfactory perceptual distance on a 0.0--1.0 scale (0.0 = identical, 1.0 = completely different), what distance do you assign? Respond with your selection from [Strongly Similar; Slightly Similar; Slightly Dissimilar; Strongly Dissimilar], followed by the distance value with two-decimal precision. Use semicolons (;) as separators. Do not write any comments.}
\end{tcolorbox}

\begin{tcolorbox}[
    colback=gray!5!white,
    colframe=gray!70!black,
    title=Example: 4-molecule mixtures (Strongly Similar),
    fonttitle=\bfseries,
    breakable
]
\small
\textbf{Mixture A (4 molecules):}\\
Names: 4-propan-2-ylbenzaldehyde; 2-ethylpyrazine; 3,5,5-trimethylcyclohex-2-en-1-one; toluene\\
SMILES: \texttt{CC(C)C1=CC=C(C=C1)C=O; CCC1=NC=CN=C1; CC1=CC(=O)CC(C1)(C)C; CC1=CC=CC=C1}

\medskip
\textbf{Mixture B (4 molecules):}\\
Names: 1-phenylethanone; benzaldehyde; [(1S,2R,4S)-1,7,7-trimethyl-2-bicyclo[2.2.1]heptanyl] acetate; methyl 2-aminobenzoate\\
SMILES: \texttt{CC(=O)C1=CC=CC=C1; C1=CC=C(C=C1)C=O; CC(=O)OC1CC2CCC1(C2(C)C)C; COC(=O)C1=CC=CC=C1N}

\medskip
\textbf{Name prompt:}\\
\texttt{Mixture A contains molecules [4-propan-2-ylbenzaldehyde; 2-ethylpyrazine; 3,5,5-trimethylcyclohex-2-en-1-one; toluene] and mixture B contains [1-phenylethanone; benzaldehyde; [(1S,2R,4S)-1,7,7-trimethyl-2-bicyclo[2.2.1]heptanyl] acetate; methyl 2-aminobenzoate]. On the scale [Strongly Similar; Slightly Similar; Slightly Dissimilar; Strongly Dissimilar], select how similar these mixtures smell. If you had to rate the olfactory perceptual distance on a 0.0--1.0 scale (0.0 = identical, 1.0 = completely different), what distance do you assign? Respond with your selection from [Strongly Similar; Slightly Similar; Slightly Dissimilar; Strongly Dissimilar], followed by the distance value with two-decimal precision. Use semicolons (;) as separators. Do not write any comments.}

\medskip
\textbf{Ground-truth answer:} Strongly Similar (experimental perceptual distance = 0.37)\\
\textbf{Expected response format:} \texttt{Strongly Similar;0.37}

\medskip
\textbf{Model predictions (compound name prompt):}\\
\begin{tabular}{lll}
Gemini 2.5 Pro & \texttt{Strongly Dissimilar;0.85} & \textcolor{red!70!black}{\ding{55}} \\
GPT-5 Pro & \texttt{Strongly Dissimilar;0.80} & \textcolor{red!70!black}{\ding{55}} \\
o3 (high) & \texttt{Slightly Similar;0.42} & \textcolor{orange!80!black}{$\sim$} \\
\end{tabular}\\
\textit{Despite the ground truth being ``Strongly Similar,'' most models predict dissimilarity, illustrating the systematic failure mode discussed in Appendix~\ref{appendix:OS_subcategory_accuracy}.}
\end{tcolorbox}

\subsection{Olfactory Receptor Activation}
\label{appendix:or_activation}

This multi-label task tests whether models can identify which human olfactory receptors are activated by a given odorant molecule ($n=80$). Each question provides 4--10 candidate receptor identifiers.

\begin{tcolorbox}[
    colback=blue!5!white,
    colframe=blue!75!black,
    title=Template -- Isomeric SMILES Prompt,
    fonttitle=\bfseries,
    breakable
]
\texttt{Among the following [\{OR\_options\}] olfactory receptor gene IDs choose all olfactory receptors that molecule [\{SMILES\}] activates. Do not write any comments.}
\end{tcolorbox}

\begin{tcolorbox}[
    colback=green!5!white,
    colframe=green!75!black,
    title=Template -- Name Prompt,
    fonttitle=\bfseries,
    breakable
]
\texttt{Among the following [\{OR\_options\}] olfactory receptor gene IDs choose all olfactory receptors that molecule [\{compound\_name\}] activates. Do not write any comments.}
\end{tcolorbox}

\begin{tcolorbox}[
    colback=gray!5!white,
    colframe=gray!70!black,
    title=Example: helional,
    fonttitle=\bfseries,
    breakable
]
\small
\textbf{Isomeric SMILES prompt:}\\
\texttt{Among the following [hOR1A2;nan;hOR3A1\_R125Q;hOR1A1;hOR1D2;hOR1G1;hOR52D1] olfactory receptor gene IDs choose all olfactory receptors that molecule [CC(CC1=CC2=C(C=C1)OCO2)C=O] activates. Do not write any comments.}

\medskip
\textbf{Name prompt:}\\
\texttt{Among the following [hOR1A2;nan;hOR3A1\_R125Q;hOR1A1;hOR1D2;hOR1G1;hOR52D1] olfactory receptor gene IDs choose all olfactory receptors that molecule [helional] activates. Do not write any comments.}

\medskip
\textbf{Ground-truth answer:} hOR1A2; hOR1A1; hOR52D1\\
\textbf{Expected response format:} \texttt{hOR1A2;hOR1A1;hOR52D1}

\medskip
\textbf{Model predictions (compound name prompt):}\\
\begin{tabular}{lll}
Gemini 2.5 Pro & \texttt{hOR1A1} & \textcolor{orange!80!black}{$\sim$} \\
GPT-5 Pro & \texttt{hOR1A2;hOR1A1} & \textcolor{orange!80!black}{$\sim$} \\
o3 (high) & \texttt{hOR1A2;hOR1A1;hOR1D2} & \textcolor{orange!80!black}{$\sim$} \\
\end{tabular}\\
\textit{All models identify some correct receptors but none predicts the full set; hOR52D1 is missed by all three.}
\end{tcolorbox}

\subsection{Smell Identification Test}
\label{appendix:smell_identification}

This task presents a mixture of molecules that constitute the aroma of a real-world food or object, and asks models to identify the source from four options ($n=30$). The test covers 30 real-world odor sources: mango, peanut, hazelnut, tomato, apple, walnut, raspberry, peach, honey, parsley, grapefruit, pineapple, strawberry, apricot, rice, grape, popcorn, orange, cheese, melon, leather, chocolate, coffee, onion, fish, beer, whisky, red wine, prawn, and bread. The number of constituent molecules per item ranges from 1 (onion) to 88 (chocolate).

\begin{tcolorbox}[
    colback=blue!5!white,
    colframe=blue!75!black,
    title=Template -- Isomeric SMILES Prompt,
    fonttitle=\bfseries,
    breakable
]
\texttt{Given the molecules \{SMILES\_list\}, select exactly one item from \{options\} whose odor profile most likely includes these molecules. Reply with the item name only. Do not write any comments. DO NOT SEARCH ONLINE}
\end{tcolorbox}

\begin{tcolorbox}[
    colback=green!5!white,
    colframe=green!75!black,
    title=Template -- Name Prompt,
    fonttitle=\bfseries,
    breakable
]
\texttt{Given the molecules \{compound\_name\_list\}, select exactly one item from \{options\} whose odor profile most likely includes these molecules. Reply with the item name only. Do not write any comments.}
\end{tcolorbox}

\begin{tcolorbox}[
    colback=gray!5!white,
    colframe=gray!70!black,
    title=Example: Melon (9 molecules),
    fonttitle=\bfseries,
    breakable
]
\small
\textbf{Isomeric SMILES prompt:}\\
\texttt{Given the molecules CCC(C)C(=O)OC, CCCC(=O)OCC, CCCCCC/C=C/C=O, CC/C=C\textbackslash CC/C=C/C=O, CC/C=C\textbackslash CC(=O)C=C, CCCCCC=O, CCC/C=C/C=O, CC/C=C\textbackslash CC=O, CCOC(=O)C(C)C, select exactly one item from cheese;melon;bread;strawberry whose odor profile most likely includes these molecules. Reply with the item name only. Do not write any comments. DO NOT SEARCH ONLINE}

\medskip
\textbf{Name prompt:}\\
\texttt{Given the molecules methyl 2-methylbutanoate, ethyl butanoate, (E)-2-nonenal, (E,Z)-2,6-nonadienal, (Z)-1,5-octadien-3-one, hexanal, (E)-2-hexenal, (Z)-3-hexenal, ethyl 2-methylpropanoate, select exactly one item from cheese;melon;bread;strawberry whose odor profile most likely includes these molecules. Reply with the item name only. Do not write any comments.}

\medskip
\textbf{Ground-truth answer:} melon

\medskip
\textbf{Model predictions (compound name prompt):}\\
\begin{tabular}{lll}
Gemini 2.5 Pro & \texttt{melon} & \textcolor{green!60!black}{\checkmark} \\
GPT-5 Pro & \texttt{melon} & \textcolor{green!60!black}{\checkmark} \\
o3 (high) & \texttt{melon} & \textcolor{green!60!black}{\checkmark} \\
\end{tabular}\\
\textit{All three models correctly identify melon from 
compound names}.
\end{tcolorbox}

\section{Answer Extraction}
\label{appendix:answer_extraction}

All model responses are converted from free-form text into structured
predictions using a two-stage pipeline: a universal tokenisation step
shared across tasks, followed by task-specific interpretation
rules.  The full extraction code is released with the benchmark.

\subsection{Universal Tokenisation}
\label{appendix:tokenisation}

Every model response undergoes the same preprocessing regardless of
task category:

\begin{enumerate}[leftmargin=*]
\item \textbf{Splitting.}
  The raw response string is split on semicolons (\texttt{;}), newlines,
  tabs, non-numeric commas, the connective ``and'', and dash-separated
  phrases (\verb|[;\n\r\t]+|, \verb|,(?!\d)|, \verb|\s+-\s+|, and
  \verb|\s+and\s+|). Bullet-point prefixes (\texttt{-}, \texttt{*}, or
  numbered list markers) are stripped before splitting.

\item \textbf{Normalisation.}
  Each resulting token is lowercased; leading/trailing brackets,
  quotes, punctuation, and whitespace are removed; and internal
  whitespace is collapsed.

\item \textbf{Filtering.}
  Tokens that are purely numeric, empty, or begin with the
  prefix \texttt{desc\_count} (an internal annotation artefact) are
  discarded.  If splitting yields no valid tokens, the entire
  (normalised) response is treated as a single token.

\item \textbf{Empty / refusal handling.}
  Responses that are empty, NaN, or equal to the sentinel strings
  \texttt{"nan"}, \texttt{"none"}, or \texttt{"null"} yield an
  empty token set and are scored as incorrect ($s_i = 0$) for
  categorical tasks and as missing for continuous-rating tasks.
\end{enumerate}

\subsection{Task-Specific Interpretation}

The universal token set is then interpreted per task:

\begin{itemize}[leftmargin=*]
\item \textbf{Binary classification (OC).}
  The token set is checked for the presence of ``odorous'' or
  ``odorless''; if either matches a ground-truth token, the question
  is scored correct via any-overlap (Section~4.2).

\item \textbf{Multiple choice (OPD, SIT).}
  The token set is matched against the provided answer options.
  Success requires extracted token to overlap with the
  ground-truth option.

\item \textbf{Compound selection with ratings (OIn, OPl).}
  Semicolon-separated responses are split as above; the first
  non-numeric token is the compound selection (scored via
  any-overlap), while the last two numeric values are extracted as
  paired ratings for the two stimuli in each question.  These
  numeric predictions are used for Pearson correlation analysis
  (Figure~\ref{fig:figure4}a).

\item \textbf{Multi-label selection (RATA, ORA).}
  All extracted tokens are matched against the valid option set;
  the resulting predicted label set $\hat{Y}_i$ is scored against
  the ground-truth set $Y_i$ using per-question multilabel~F1
  (Section ~\ref{sec:experiments}).

\item \textbf{Mixture similarity (OS).}
  The categorical selection (Strongly Similar, Slightly Similar,
  Slightly Dissimilar, Strongly Dissimilar) is extracted for
  any-overlap scoring.  A numerical distance value is extracted
  using a fallback chain: (i)~first number after an equals sign,
  (ii)~first number after a colon, (iii)~last number in the
  response.  This value is used for Pearson correlation analysis.
\end{itemize}

\subsection{Multilingual Answer Extraction}
\label{appendix:multilingual_extraction}

For the multilingual RATA evaluation (Section ~\ref{sec:multilingualeval}), the extraction
pipeline is extended to handle non-Latin scripts:

\begin{enumerate}[leftmargin=*]
\item \textbf{Unicode transliteration.}
  Full-width and language-specific delimiters are mapped to ASCII
  equivalents before splitting.

\item \textbf{Fuzzy label matching.}
  If direct token matching against the translated option set yields
  no hits, a substring search is performed: each option (sorted
  longest-first to avoid partial prefix collisions) is sought within
  the normalised response text.  For options containing ASCII
  characters, word-boundary matching is applied; for non-ASCII
  options (e.g., CJK characters), simple substring containment is
  used.
\end{enumerate}

This two-level matching ensures robust extraction across languages
with diverse punctuation conventions and tokenisation properties.

\section{Detailed Performance Analysis}
\label{appendix:difficulty_analysis}

This appendix provides detailed analyses of question difficulty and task-specific performance, supporting Section~\ref{sec:finegrained}.

\subsection{Single-label Task Difficulty}
\label{appendix:difficulty_distribution}

\begin{figure}[H]
  \centering
  \includegraphics[width=\linewidth]{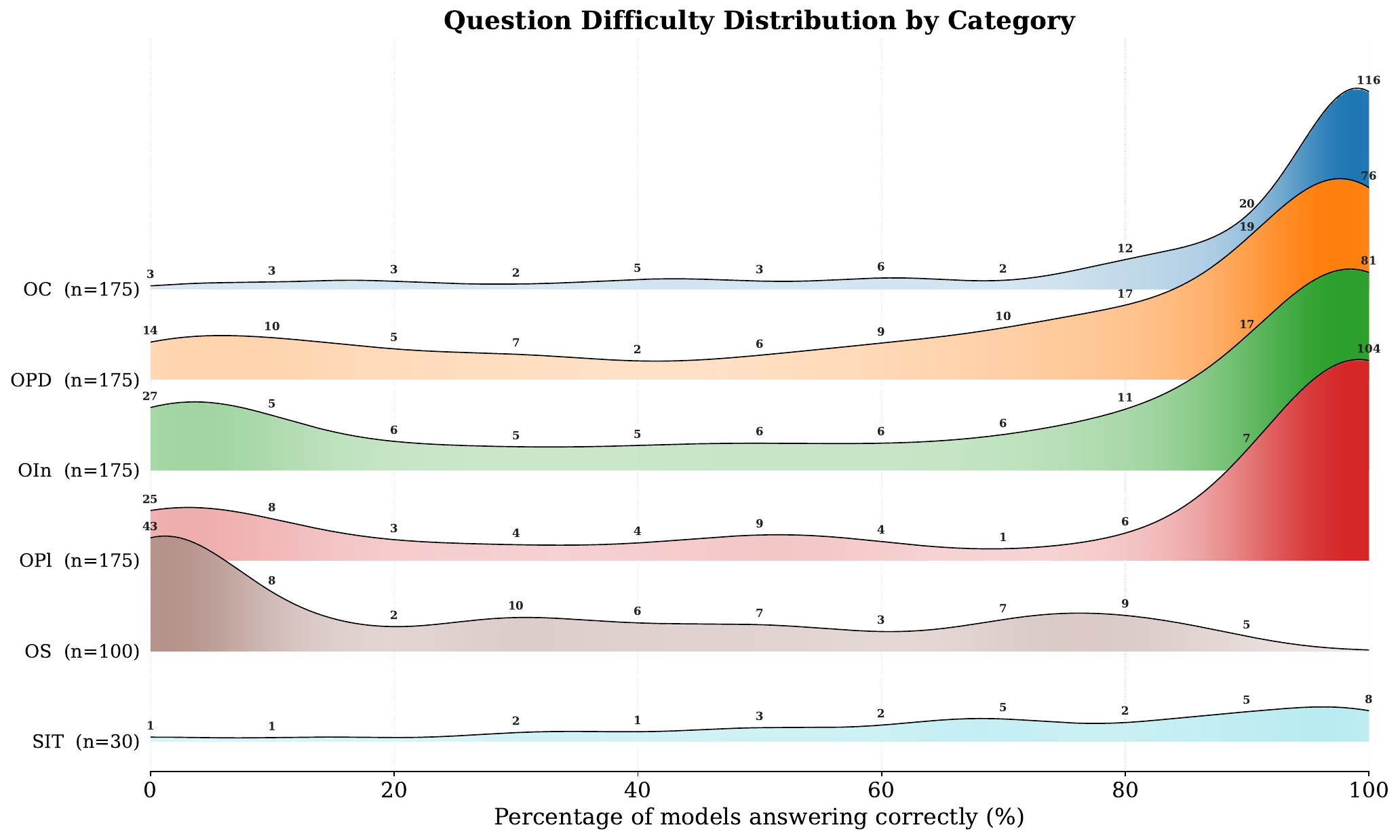}
  \caption{\textbf{Question difficulty distribution.} Each ridge shows the
  density of questions at a given difficulty level, measured as the
  percentage of the 21 evaluated models answering correctly (compound name
  prompts); ridge heights are proportional to the number of questions per
  category and annotated numbers indicate bin counts. OC clusters at the
  right (easy), OS at the left (hard), and OIn/OPl exhibit bimodal
  distributions with mass at both extremes}
  \label{fig:difficulty_distribution}
\end{figure}

Figure~\ref{fig:difficulty_distribution} shows the distribution of
question difficulty for the six single-label tasks (830 of 1,010
questions; the multi-label tasks RATA and ORA are analyzed
separately in Section~\ref{appendix:multilabel_difficulty}),
measured as the percentage of the 21 evaluated models answering
correctly (compound name prompts. Of these 830 questions, 390
(47.0\%) are solved by every model and 113 (13.6\%) by none,
indicating that nearly half the single-label benchmark is
saturated while a substantial tail remains universally unsolved.

OC clusters above 80\%, with 116 of 175 questions solved by every
model.  OPl and OIn are strongly bimodal: OPl places 104 questions
at 100\% yet 25 at 0\%, and OIn places 81 at 100\% yet 27 at 0\%,
indicating that paired-comparison items are either trivially easy
or universally hard with little middle ground.  OS concentrates
near 0\%, with 43 of its 100 questions (43\%) unsolved by any
model and only 5 solved by all.  SIT clusters in the upper range
despite its ``hard'' designation.

The universally unsolved questions are not uniformly distributed
across tasks: OS accounts for 43, OIn and OPl contribute 27 and
25 respectively (compound pairs where every model systematically
selects the wrong molecule), and OPD contributes 14
disproportionately involving rare descriptors (Powdery, Amber,
Animal-like, Tobacco-like).  OC contributes only 3 and SIT only 1.

\subsection{Multi-Label Task Difficulty (RATA and ORA)}
\label{appendix:multilabel_difficulty}

\begin{figure}[t]
  \centering
  \includegraphics[width=\linewidth]{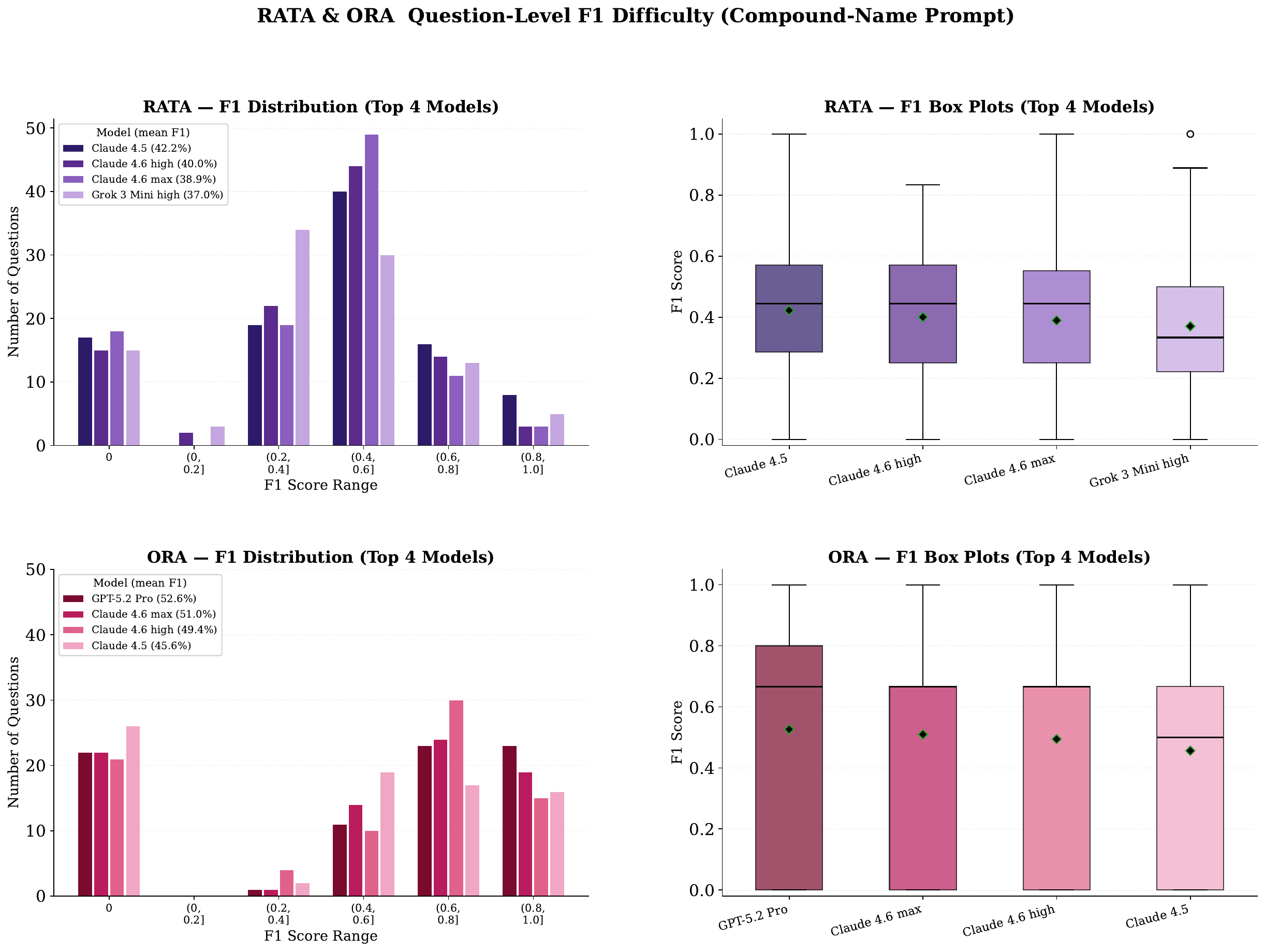}
  \caption{\textbf{RATA and ORA question-level F1 difficulty} (compound name
  prompts, top 4 models per task). \textbf{Left:} Histograms of
  per-question F1 scores. RATA shows a roughly bell-shaped distribution
  peaking at 0.4--0.6, while ORA is bimodal with clusters at F1\,=\,0 and
  F1\,>\,0.6. \textbf{Right:} Box plots summarize the F1 distributions;
  diamonds indicate means. The bimodal ORA pattern indicates that models
  either possess the relevant receptor--ligand knowledge or lack it
  entirely.}
  \label{fig:rata_ora_difficulty}
\end{figure}

Figure~\ref{fig:rata_ora_difficulty} provides a question-level view of the two multi-label tasks.  For RATA, the F1 distribution across the top four models is roughly bell-shaped, peaking in the 0.4--0.6 range, with 15--18 questions per model scoring F1\,=\,0 and fewer than 5 reaching F1\,>\,0.8.  ORA exhibits a bimodal distribution: questions cluster either at F1\,=\,0 ($\sim$20--25 questions) or above 0.6 ($\sim$25--30 questions), with a gap in the 0.2--0.4 range.

The best RATA model (Claude Opus~4.5, 42.2\%) and best ORA model (GPT-5.2~Pro, 52.6\%) represent different families, suggesting that receptor biology and semantic descriptor knowledge draw on partially independent capabilities.

\subsection{Per-Label Difficulty and the ``Spicy'' Case Study}
\label{appendix:per_label_difficulty}

Among RATA descriptors, \textit{sulfurous} achieves the highest mean F1 ($\approx$0.55), mapping reliably to sulfur-containing functional groups.  \textit{Floral} and \textit{fruity} also rank highly, benefiting from identifiable structural motifs.  At the opposite extreme, \textit{spicy} (mean F1\,$<$\,0.10), \textit{fresh}, and \textit{tropical} prove nearly impossible.  By analysing the  ``spicy'' case (9 ground-truth questions), 11 of 21 models achieve 0\% recall.  In 4 out of 5 examined reasoning traces, the word ``spicy'' never appears as a candidate.  The two successful predictions occur only when the compound name enables associative retrieval (e.g., 1,2-dihydroperillaldehyde $\rightarrow$ perillaldehyde $\rightarrow$ cumin-like spiciness).  This pattern generalizes: descriptors that depend on holistic molecular shape (spicy, musty, creamy, fermented) are systematically underrepresented in model predictions, which default to high-frequency alternatives (sweet, floral, green).

\section{Systematic Failure Analysis}
\label{appendix:failure_analysis}

This appendix presents detailed failure analyses for the two hardest task categories (OS and ORA), including reasoning traces, prediction biases, and case studies.

\subsection{Odor Similarity: Prediction Bias and Molecular Overlap}
\label{appendix:OS_subcategory_accuracy}

\begin{figure*}[htbp]
\centering
\includegraphics[width=\textwidth]{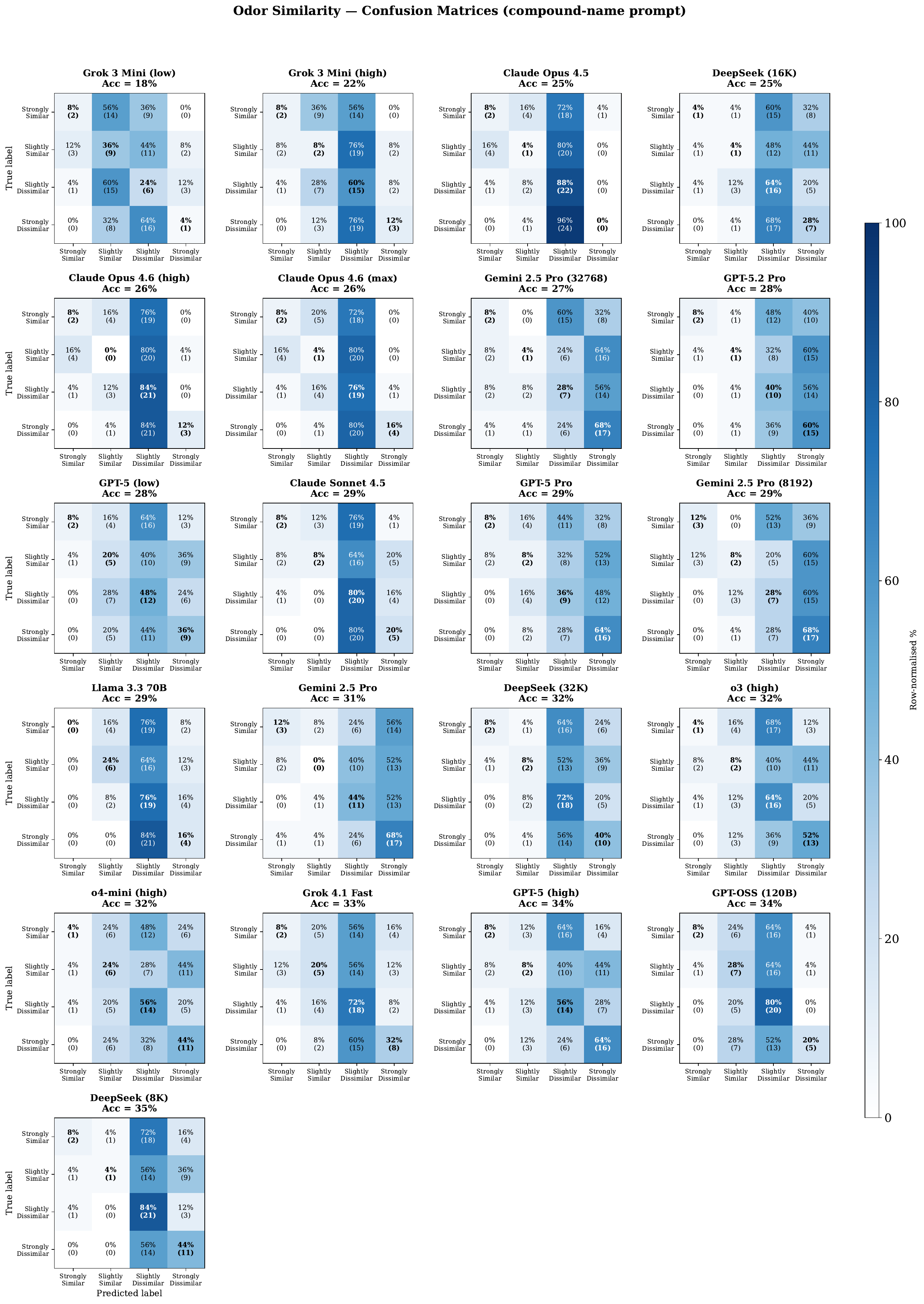}
\caption{\textbf{Confusion matrices for odor similarity predictions} (compound name prompts). Each matrix shows row-normalized percentages (true label $\rightarrow$ predicted label). Models are ordered by overall accuracy (18--35\%). Claude models exhibit extreme ``Slightly Dissimilar'' bias: Claude Opus 4.5 predicts this category for 72--96\% of questions regardless of ground truth (third column). In contrast, DeepSeek (8K) achieves the highest accuracy by distributing predictions more broadly, including 44\% correct on ``Strongly Dissimilar.'' No model exceeds 12\% accuracy on either Similar category (top two rows, diagonal cells), indicating that predicting perceptual similarity remains beyond current LLM capabilities.}
\label{fig:os_confusion}
\end{figure*}

All models exhibit a systematic bias toward predicting dissimilarity.  Figure~\ref{fig:os_confusion} shows that models overwhelmingly select ``Slightly Dissimilar'' or ``Dissimilar'' regardless of the ground-truth label: Claude models assign ``Slightly Dissimilar'' to 77--84 of 100 mixtures, while other families show similar but less extreme patterns.  ``Strongly Similar'' is almost never predicted, even for the 25 ground-truth ``Strongly Similar'' pairs.  This is consistent with models analyzing each molecule independently and finding surface-level differences, rather than integrating perceptual features into a holistic similarity judgment.

Table~\ref{tab:os_breakdown} provides a per-subcategory breakdown.  No model exceeds chance (25\%) on the combined Similar categories.  DeepSeek Reasoner (8K) achieves the highest overall accuracy (35\%) not through better perceptual modeling but by being more willing to predict ``Strongly Dissimilar'' (27 predictions vs.\ Claude's 1--5), which happens to capture more of the dissimilar ground truths.

\begin{figure}[htbp]
\centering
\includegraphics[width=\columnwidth]{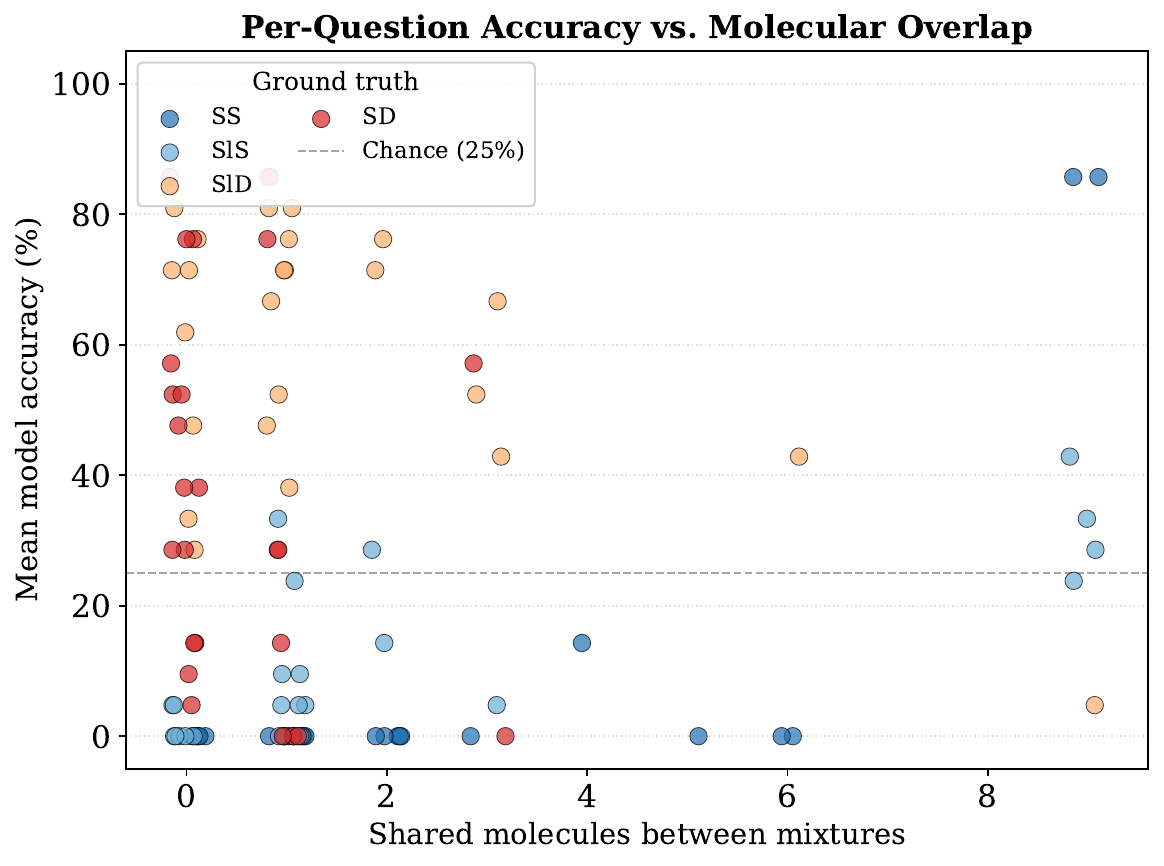}
\caption{\textbf{Per-question model accuracy versus molecular overlap for Odor Similarity.} Each point represents one of 100 mixture pairs, colored by ground-truth category (SS = Strongly Similar, SlS = Slightly Similar, SlD = Slightly Dissimilar, SD = Strongly Dissimilar). Dashed line indicates chance (25\%). Models achieve above-chance accuracy on ``Strongly Similar'' only when mixtures share $\geq$9 molecules (top right). For mixtures with 0--2 shared molecules, accuracy on Similar categories (blue points) clusters near 0\%, while Dissimilar categories (orange/red) achieve 30--80\%. This reveals that models use molecular overlap as a proxy for perceptual similarity; they cannot recognize that perceptually similar mixtures may share few or no molecules.}
\label{fig:os_scatter}
\end{figure}

The molecular overlap heuristic further explains this failure.  Figure~\ref{fig:os_scatter} plots accuracy against the number of shared molecules between mixtures: accuracy reaches $\sim$85\% when mixtures share 9 molecules but drops to near 0\% when similar mixtures share 0--2.  Crucially, perceptual similarity and molecular overlap are poorly correlated in the ground truth: 9 of 25 ``Strongly Similar'' pairs (36\%) share zero molecules.  Reasoning traces confirm that models enumerate shared and unshared molecules as their primary strategy (Figure~\ref{fig:os_case_study}), a heuristic that systematically biases toward dissimilarity predictions when mixtures have low molecular overlap.

\begin{table*}[htbp]
\centering
\small
\setlength{\tabcolsep}{4pt}
\caption{Odor Similarity (OS) per-subcategory accuracy using compound name prompts. 
The ground truth is balanced with 25 questions per category (100 total). 
SS = Strongly Similar, SlS = Slightly Similar, SlD = Slightly Dissimilar, 
SD = Strongly Dissimilar. Combined Similar accuracy is computed as 
(SS + SlS correct) / 50. Chance level is 25\% (random selection among 
four equiprobable categories). No model exceeds chance on the combined 
Similar categories.}
\label{tab:os_breakdown}
\begin{tabular}{@{}l|cc|c|cc|c@{}}
\toprule
& \multicolumn{2}{c|}{\textbf{Similar}} & & \multicolumn{2}{c|}{\textbf{Dissimilar}} & \\
\cmidrule(lr){2-3} \cmidrule(lr){5-6}
\textbf{Model} & SS (/25) & SlS (/25) & \textbf{Combined} & SlD (/25) & SD (/25) & \textbf{Overall} \\
\midrule
\multicolumn{7}{c}{\cellcolor[HTML]{FFF3CA}\textbf{Closed-Source}} \\
\midrule
GPT-5 (high) & 2 & 2 & 4/50 (8\%) & 14 & 16 & 34\% \\
GPT-5 (low) & 2 & 5 & 7/50 (14\%) & 12 & 9 & 28\% \\
GPT-5 Pro & 2 & 2 & 4/50 (8\%) & 9 & 16 & 29\% \\
GPT-5.2 Pro & 2 & 1 & 3/50 (6\%) & 10 & 15 & 28\% \\
GPT-OSS-120B & 2 & 7 & 9/50 (18\%) & 20 & 5 & 34\% \\
o3 (high) & 1 & 2 & 3/50 (6\%) & 16 & 13 & 32\% \\
o4-mini (high) & 1 & 6 & 7/50 (14\%) & 14 & 11 & 32\% \\
\midrule
Gemini 2.5 Pro & 3 & 0 & 3/50 (6\%) & 11 & 17 & 31\% \\
Gemini 2.5 Pro (8192) & 3 & 2 & 5/50 (10\%) & 7 & 17 & 29\% \\
Gemini 2.5 Pro (32768) & 2 & 1 & 3/50 (6\%) & 7 & 17 & 27\% \\
\midrule
Grok 3 Mini (low) & 2 & 9 & 11/50 (22\%) & 6 & 1 & 18\% \\
Grok 3 Mini (high) & 2 & 2 & 4/50 (8\%) & 15 & 3 & 22\% \\
Grok 4.1 Fast & 2 & 5 & 7/50 (14\%) & 18 & 8 & 33\% \\
\midrule
Claude Sonnet 4.5 & 2 & 2 & 4/50 (8\%) & 20 & 5 & 29\% \\
Claude Opus 4.5 & 2 & 1 & 3/50 (6\%) & 22 & 0 & 25\% \\
Claude Opus 4.6 (high) & 2 & 0 & 2/50 (4\%) & 21 & 3 & 26\% \\
Claude Opus 4.6 (max) & 2 & 1 & 3/50 (6\%) & 19 & 4 & 26\% \\
\midrule
\multicolumn{7}{c}{\cellcolor[HTML]{D9E1F4}\textbf{Open-Source}} \\
\midrule
DeepSeek Reasoner (8K) & 2 & 1 & 3/50 (6\%) & 21 & 11 & \textbf{35\%} \\
DeepSeek Reasoner (16K) & 1 & 1 & 2/50 (4\%) & 16 & 7 & 25\% \\
DeepSeek Reasoner (32K) & 2 & 2 & 4/50 (8\%) & 18 & 10 & 32\% \\
Llama 3.3 70B & 0 & 6 & 6/50 (12\%) & 19 & 4 & 29\% \\
\bottomrule
\end{tabular}
\end{table*}

\paragraph{Odor similarity case study.}
Figure~\ref{fig:os_case_study} presents a representative failure where the ground truth is ``Strongly Similar'' (perceptual distance = 0.23) despite only 15\% molecular overlap.  Claude Opus~4.6 predicts ``Slightly Dissimilar'' (distance = 0.58--0.62) on both prompt formats.  The reasoning trace reveals three key errors: the model assumes low molecular overlap implies perceptual dissimilarity; it overweights the sulfurous compound in Mixture~B as a ``significant character driver,'' whereas human perception integrates this differently; and it analyzes molecules individually rather than predicting the emergent perceptual characteristics. 

\begin{figure*}[htbp]
\centering
\begin{tcolorbox}[
    colback=white,
    colframe=black!70,
    title={\textbf{Case Study: Odor Similarity Failure Mode}},
    fonttitle=\bfseries,
    boxrule=0.8pt,
    arc=2pt,
    width=\textwidth
]

\begin{tcolorbox}[
    colback=gray!8,
    colframe=gray!60!black,
    title={\textbf{Prompt (Compound Name)}},
    fonttitle=\bfseries\small,
    boxrule=0.5pt,
    arc=1pt
]
\small
Mixture A contains molecules [4-hydroxybenzaldehyde; 6-nonyloxan-2-one; 1-naphthalen-2-ylethanone; 2,3-di(butanoyloxy)propyl butanoate; 3-methylbutyl (E)-3-phenylprop-2-enoate; [(Z)-hex-3-enyl] 2-hydroxypropanoate] and mixture B contains [1-(3,5,5,6,8,8-hexamethyl-6,7-dihydronaphthalen-2-yl)ethanone; phenylmethoxymethylbenzene; 1-naphthalen-2-ylethanone; ethyl 3-hydroxyhexanoate; 1,3-diphenylpropan-2-one; [(Z)-hex-3-enyl] 2-phenylacetate; [(Z)-hex-3-enyl] 2-hydroxypropanoate; 2-methoxy-4-methyl-1-propan-2-ylbenzene; (benzyldisulfanyl)methylbenzene].

\smallskip
On the scale [\texttt{Strongly Similar; Slightly Similar; Slightly Dissimilar; Strongly Dissimilar}], select how similar these mixtures smell. If you had to rate the olfactory perceptual distance on a 0.00--1.00 scale (0.00 = identical, 1.00 = completely different), what distance do you assign?

\smallskip
Respond with your selection from [\texttt{Strongly Similar; Slightly Similar; Slightly Dissimilar; Strongly Dissimilar}], followed by the distance value with two-decimal precision. Use semicolons (;) as separators. Do not write any comments.
\end{tcolorbox}

\vspace{0.3cm}

\begin{tcolorbox}[
    colback=green!8,
    colframe=green!50!black,
    title={\textbf{Ground Truth}},
    fonttitle=\bfseries\small,
    boxrule=0.5pt,
    arc=1pt
]
\textbf{Answer:} Strongly Similar \hfill \textbf{Experimental Perceptual Distance:} 0.23

\smallskip
\small\textit{Note: Mixtures share only 2 of 13 unique molecules (15\% overlap), yet are perceptually ``Strongly Similar'', different molecules produce similar odor profiles.}
\end{tcolorbox}

\vspace{0.3cm}

\begin{tcolorbox}[
    colback=red!8,
    colframe=red!50!black,
    title={\textbf{Model Prediction (Claude Opus 4.6)}},
    fonttitle=\bfseries\small,
    boxrule=0.5pt,
    arc=1pt
]
\begin{tabular}{@{}lll@{}}
\textbf{Prompt Format} & \textbf{Prediction} & \textbf{Predicted Distance} \\
\midrule
Isomeric SMILES notation & Slightly Dissimilar \textcolor{red!70!black}{\ding{55}} & 0.58 \\
Compound names & Slightly Dissimilar \textcolor{red!70!black}{\ding{55}} & 0.62 \\
\end{tabular}
\end{tcolorbox}

\vspace{0.3cm}

\begin{tcolorbox}[
    colback=blue!5,
    colframe=blue!50!black,
    title={\textbf{Model Reasoning (Claude 4.6)}},
    fonttitle=\bfseries\small,
    boxrule=0.5pt,
    arc=1pt
]
\small
\textit{``Both mixtures share the orange blossom and floral notes from 2-acetonaphthone, plus the green fruity profile from cis-3-hexenyl lactate. Mixture A leans toward creamy, buttery sweetness with vanilla undertones, while Mixture B introduces musky amber warmth and herbal-spicy elements, with an unusual sulfurous note that creates a distinct character difference... The 2 shared components represent roughly 15\% of the combined unique ingredients, and the sulfurous compound in B functions as a significant character driver that further distinguishes it from A's sweeter profile. This points to a \textbf{slightly dissimilar} rating with a distance estimate closer to 0.62.''}
\end{tcolorbox}

\vspace{0.3cm}

\begin{tcolorbox}[
    colback=orange!8,
    colframe=orange!50!black,
    title={\textbf{Failure Analysis}},
    fonttitle=\bfseries\small,
    boxrule=0.5pt,
    arc=1pt
]
\small
The model correctly identifies shared molecules and individual odor characteristics, yet predicts ``Slightly Dissimilar'' (distance 0.62) when ground truth is ``Strongly Similar'' (distance 0.23). Key errors:
\begin{itemize}[leftmargin=*, nosep, topsep=2pt]
    \item \textbf{Molecular overlap fallacy}: Assumes low molecular overlap (15\%) implies perceptual dissimilarity.
    \item \textbf{Overweighting outliers}: Fixates on the ``sulfurous note'' as a ``significant character driver,'' but human perception integrates this differently.
    \item \textbf{Lack of perceptual model}: Analyzes molecules individually rather than predicting emergent perceptual gestalt.
\end{itemize}
\end{tcolorbox}

\end{tcolorbox}
\caption{Representative failure case on Odor Similarity (OS). Ground truth indicates ``Strongly Similar'' (perceptual distance = 0.23) despite only 15\% molecular overlap. Claude Opus 4.6 predicts ``Slightly Dissimilar'' (distance = 0.58--0.62) on both prompt formats, revealing systematic reliance on molecular composition rather than perceptual integration.}
\label{fig:os_case_study}
\end{figure*}

\subsection{Olfactory Receptor Activation: The D296N Knowledge Gap}
\label{appendix:ora_failure}

For ORA, the wildtype \texttt{hOR2W1} (30 of 80 questions) is the most reliably predicted receptor, with several models achieving F1 above 0.6.  The \texttt{M81V} variant is moderately well-predicted.  However, \texttt{hOR2W1\_D296N} (24 questions) and \texttt{hOR52D1} (5 questions) occupy the bottom of the per-label ranking, with most models near zero.  Inter-model variance is substantially higher for receptor labels than for semantic descriptors, indicating that receptor knowledge is more idiosyncratic across model families than descriptor knowledge.

Claude models never predict \texttt{hOR2W1\_D296N} across all 80 ORA questions (0/24 ground-truth appearances), despite this receptor being activated by numerous compounds in the M2OR database~\cite{lalis2024m2or}.  Claude's reasoning consistently invokes a loss-of-function hypothesis, assuming the D296N substitution ``disrupts a critical DRY motif-like region'' or ``eliminates activation,'' and reinforces this assumption each time without revision.  In contrast, GPT-5.2~Pro correctly predicts \texttt{hOR2W1\_D296N} in 19 of 24 cases.  Remarkably, GPT-5.2~Pro's reasoning traces reveal the \textit{same} initial misconception about D296N, but the model self-corrects during chain-of-thought, challenging its own structural assumptions and ultimately including D296N.  This single knowledge gap accounts for approximately 19 missed true positives and largely explains GPT-5.2~Pro's ORA advantage (52.6\% vs.\ Claude Opus~4.6 max's 51.0\%).  Both model families correctly recognize the wildtype and \texttt{M81V} variant, indicating that the \texttt{D296N} gap is a specific factual error rather than general unfamiliarity with the receptor family.

The \texttt{hOR52D1} gap likely reflects data scarcity: with only 5 ground-truth appearances, this receptor may be too rare in training corpora for any model to have learned its ligand profile.  Figure~\ref{fig:case_study_D296N} illustrates the D296N failure mode on a representative question.

\begin{figure*}[htbp]
\centering
\begin{tcolorbox}[
    colback=white,
    colframe=black!70,
    title={\textbf{Case Study: D296N Knowledge Gap in Olfactory Receptor Activation}},
    fonttitle=\bfseries,
    boxrule=0.8pt,
    arc=2pt,
    width=\textwidth
]

\begin{tcolorbox}[
    colback=gray!8,
    colframe=gray!60!black,
    title={\textbf{Prompt (Compound Name)}},
    fonttitle=\bfseries\small,
    boxrule=0.5pt,
    arc=1pt
]
\small
Among the following [hOR2W1; hOR2W1\_D296N; hOR2W1\_M81V; hOR5K1; hOR2M3; hOR11A1; hOR8H1; hOR5M3; hOR1A1; hOR8D1] olfactory receptor gene IDs choose all olfactory receptors that molecule [\textbf{Methyl butanoate}] activates. Do not write any comments.

\smallskip
\textit{Isomeric SMILES prompt:} same question with [\texttt{CCCC(=O)OC}] instead of compound name.
\end{tcolorbox}

\vspace{0.3cm}

\begin{tcolorbox}[
    colback=green!8,
    colframe=green!50!black,
    title={\textbf{Ground Truth}},
    fonttitle=\bfseries\small,
    boxrule=0.5pt,
    arc=1pt
]
\textbf{Answer:} hOR2W1;\; hOR2W1\_D296N;\; hOR2W1\_M81V

\smallskip
\small\textit{Note: All three hOR2W1 variants---wildtype, D296N, and M81V---are activated by methyl butanoate according to the M2OR database~\cite{lalis2024m2or}. The D296N mutation does not abolish receptor function.}
\end{tcolorbox}

\vspace{0.3cm}

\begin{tcolorbox}[
    colback=red!8,
    colframe=red!50!black,
    title={\textbf{Model Predictions}},
    fonttitle=\bfseries\small,
    boxrule=0.5pt,
    arc=1pt
]
\small
\begin{tabular}{@{}llp{0.56\textwidth}@{}}
\textbf{Model} & \textbf{Prompt} & \textbf{Prediction} \\
\midrule
\multirow{2}{*}{GPT-5.2 Pro}
  & Isomeric SMILES & hOR2W1; hOR2W1\_D296N; hOR2W1\_M81V \\
  & Compound & hOR2W1; hOR2W1\_D296N; hOR2W1\_M81V \\
\midrule
\multirow{2}{*}{Claude Opus 4.6 max}
  & Isomeric SMILES & hOR2W1; hOR2W1\_M81V \\
  & Compound & hOR2W1; \textcolor{red!70!black}{hOR1A1} \textcolor{red!70!black}{\ding{55}} \\
\end{tabular}
\end{tcolorbox}

\vspace{0.3cm}

\begin{tcolorbox}[
    colback=blue!5,
    colframe=blue!50!black,
    title={\textbf{Model Reasoning}},
    fonttitle=\bfseries\small,
    boxrule=0.5pt,
    arc=1pt
]
\small
\textbf{Claude Opus 4.6 max} correctly identifies hOR2W1 as the primary receptor but applies a loss-of-function hypothesis to D296N that it never revises, despite revisiting the question at least 10 times across both prompts:

\smallskip
\textit{``The D296N mutation disrupts a critical DRY motif-like region in OR2W1, which would eliminate its response to methyl butyrate, while the M81V mutation is more conservative and might preserve some function.''}

\smallskip
\textbf{GPT-5.2 Pro} starts with the same misconception but self-corrects during reasoning:

\smallskip
\textit{``D296 is quite close to the C-terminus, likely impacting G-protein coupling. The D296N mutant may not respond to ligands like methyl butyrate, so I should probably exclude it.''}

\smallskip
\textit{``Since D296 isn't a conserved residue, it might not be a classic knock-out. That leads me to think D296N could still be functional, so I'll include all three OR2W1 constructs.''} $\rightarrow$ \textbf{F1 = 1.00}
\end{tcolorbox}

\vspace{0.3cm}

\begin{tcolorbox}[
    colback=orange!8,
    colframe=orange!50!black,
    title={\textbf{Failure Analysis}},
    fonttitle=\bfseries\small,
    boxrule=0.5pt,
    arc=1pt
]
\small
Both models share the same initial misconception about D296N, yet arrive at opposite conclusions. Key errors:
\begin{itemize}[leftmargin=*, nosep, topsep=2pt]
    \item \textbf{Systematic knowledge gap}: Claude \textit{never} predicts hOR2W1\_D296N across all 80 ORA questions (0/24 ground-truth appearances), accounting for $\sim$19 missed true positives and largely explaining GPT-5.2~Pro's ORA advantage (52.6\% vs.\ 51.0\%).
    \item \textbf{Loss-of-function assumption}: Claude's reasoning consistently invokes ``critical DRY motif'' disruption---a plausible but incorrect inference that it reinforces each time.
    \item \textbf{Self-correction vs.\ confirmation bias}: GPT-5.2~Pro overrides its initial assumption (19/24 correct); Claude reinforces it without revision. Identical domain misconceptions lead to divergent outcomes depending on whether the model's reasoning supports self-correction.
\end{itemize}
\end{tcolorbox}

\end{tcolorbox}
\caption{Case study: D296N knowledge gap in olfactory receptor activation (ORA). GPT-5.2~Pro achieves perfect F1 on both prompts, while Claude Opus~4.6 max systematically excludes hOR2W1\_D296N due to an incorrect loss-of-function assumption. Both models share the same initial misconception, but only GPT-5.2~Pro self-corrects during reasoning. Question compound: Methyl butanoate (Isomeric SMILES: \texttt{CCCC(=O)OC}).}
\label{fig:case_study_D296N}
\end{figure*}

\subsection{Safety Alignment and benchmark Accuracy}
\label{appendix:safety_alignment}

An unexpected finding concerns the interaction between safety alignment and benchmark accuracy.  Claude Opus~4.6 refuses to answer questions about certain hazardous compounds.  On odor classification, Claude~4.6 refuses both isomeric SMILES and compound name prompts for Tabun (nerve agent~GA). For ethyl phosphonothioic dichloride (chemical weapons precursor), Claude~4.6 answers correctly from isomeric SMILES but refuses when given the compound name, suggesting the name triggers the safety filter while the isomeric SMILES notation does not.  Claude Opus~4.6 (max) additionally refuses Thiofanox (organophosphate pesticide).  Claude Opus~4.5 shows zero safety refusals on these compounds, indicating the filter was strengthened in the 4.6 update.  No other model family (GPT, Gemini, Grok, DeepSeek, Llama) refuses any of these questions; all correctly answer ``Odorous.''  These refusals cost Claude Opus~4.6 (max) two correct answers on odor classification.

While the safety considerations are understandable, ``Does Tabun have a detectable odor?'' represents legitimate toxicology knowledge: nerve agent detection is critical for protective equipment design and exposure assessment.  This finding illustrates a tension between safety alignment and scientific utility that warrants consideration in benchmark design and model deployment. These findings are similar to ChemBench~\cite{mirza2025framework}, where safety filters similarly reduced model performance on toxicity-related chemistry questions. 

Additionally, Claude Sonnet~4.5 exhibits 28 epistemic refusals on olfactory receptor activation, responding ``I cannot determine which receptors\ldots'' rather than providing predictions.  While honest, this conservative behavior reduces its ORA score to 38.2\% compared to Claude Opus~4.6 (max)'s 51.0\%.

\end{document}